%% file: 0-main.tex
\documentclass[sigconf]{acmart}

\usepackage{amsmath}
\usepackage{bm}
\usepackage{mathrsfs}
\usepackage{amsthm}
\usepackage{amsfonts}
\usepackage{subfigure}
\usepackage{multirow}
\usepackage{url}
\usepackage{array}
\usepackage{colortbl}
\usepackage[normalem]{ulem}
\usepackage[ruled,linesnumbered,vlined]{algorithm2e}
\usepackage{algorithmic}

\AtBeginDocument{%
  \providecommand\BibTeX{{%
    \normalfont B\kern-0.5em{\scshape i\kern-0.25em b}\kern-0.8em\TeX}}}
\setcopyright{rightsretained}
\setcopyright{acmcopyright}
\copyrightyear{2022}
\acmYear{2022}
\acmDOI{XXXXXXX.XXXXXXX}

\acmConference[CIKM'22]{the 31st ACM International Conference on Information and Knowledge Management}{October 17-22, 2022}{Hybrid Conference, Hosted in Atlanta, Georgia, USA}
\acmPrice{15.00}
\acmISBN{978-1-4503-XXXX-X/18/06}

\newcommand{\framework}{DRL-DBSCAN}
\newcommand{\tobeupdated}{\textcolor{black}}

\begin{document}
\title{Automating DBSCAN via Deep Reinforcement Learning}

\author{Ruitong Zhang$^{1}$, Hao Peng$^{1}$, Yingtong Dou$^2$, Jia Wu$^3$, Qingyun Sun$^1$, Jingyi Zhang$^1$, Philip S. Yu$^2$}
\affiliation{
\institution{
$^1$ Beihang University;
$^2$ University of Illinois Chicago;
$^3$ Macquarie University\\
$^*$ Corresponding author (penghao@buaa.edu.cn)\\
}
}

\renewcommand{\shortauthors}{Ruitong Zhang, et al.}

\theoremstyle{definition}
\newtheorem{define}{Definition}[]

\begin{abstract}
\tobeupdated{DBSCAN is widely used in many scientific and engineering fields because of its simplicity and practicality.}
\tobeupdated{However, due to its high sensitivity parameters, the accuracy of the clustering result depends heavily on practical experience.}
In this paper, we first propose a novel \textbf{D}eep \textbf{R}einforcement \textbf{L}earning guided automatic \textbf{DBSCAN} parameters search framework, namely \textbf{\framework}.
The framework models the process of adjusting the parameter search direction by perceiving the clustering environment as a Markov decision process, which aims to find the best clustering parameters without manual assistance.
{\framework} learns the optimal clustering parameter search policy for different feature distributions via interacting with the clusters, using a weakly-supervised reward training policy network.
In addition, we also present a recursive search mechanism driven by the scale of the data to efficiently and controllably process large parameter spaces.
Extensive experiments are conducted on five artificial and real-world datasets based on the proposed four working modes.
The results of offline and online tasks show that the \framework~not only consistently improves DBSCAN clustering accuracy by up to $26\%$ and $25\%$ respectively, but also can stably find the dominant parameters with high computational efficiency.
\tobeupdated{The code is available at \href{https://github.com/RingBDStack/DRL-DBSCAN}{GitHub}.}



\end{abstract}

\ccsdesc[500]{Information systems~Clustering}
\ccsdesc[300]{Computing methodologies~Artificial intelligence}

\keywords{Density-based clustering, hyperparameter search, deep reinforcement learning, recursive mechanism}

\maketitle

\input{1-intro}

\input{2-definition}

\input{3-model}

\input{4-experiment}

\input{5-relatedwork}

\input{6-conclusion}

\begin{acks}
\tiny{The authors of this paper were supported by the National Key R\&D Program of China through grant 2021YFB1714800, NSFC through grants U20B2053 and 62002007, S\&T Program of Hebei through grant 20310101D, Beijing Natural Science Foundation through grant 4222030, and the Fundamental Research Funds for the Central Universities. 
Philip S. Yu was supported by NSF under grants III-1763325, III-1909323, III-2106758, and SaTC-1930941.
Thanks for computing infrastructure provided by Huawei MindSpore platform.}
\end{acks}

\bibliographystyle{ACM-Reference-Format}
\bibliography{sample}

\newpage
\appendix
\renewcommand\thefigure{\Alph{section}\arabic{figure}} 
\renewcommand\thetable{\Alph{section}\arabic{table}} 
\renewcommand\theequation{A.\arabic{equation}}
\setcounter{table}{0}
\setcounter{figure}{0}
\setcounter{equation}{0}

\end{document}

%% file: 1-intro.tex
\section{INTRODUCTION}\label{sec:intro}

Density-Based Spatial Clustering of Applications with Noise (DBSCAN)~\cite{ester1996dbscan} is a typical density based clustering method that determines the cluster structure according to the tightness of the sample distribution.
It automatically determines the number of final clusters according to the nature of the data, has low sensitivity to abnormal points, and is compatible with any cluster shape.
In terms of application areas, benefiting from its strong adaptability to datasets of unknown distribution, DBSCAN is the preferred solution for many clustering problems, and has achieved robust performance in fields such as financial analysis~\cite{huang2019time_series,yang2014suspicious_transactions}, commercial research~\cite{fan2021consumer,wei2019milan}, urban planning~\cite{li2007public_facility,pavlis2018retail_center}, seismic research~\cite{fan2019seismic_data,kazemi2017iran,vijay2019catalogs}, recommender system~\cite{guan2018social_tagging,kuzelewska2015music}, genetic engineering~\cite{francis2011dna_damage,mohammed2018genes_patterns}, etc.

\begin{figure}[t]
  \centering
  \includegraphics[width=0.46\textwidth]{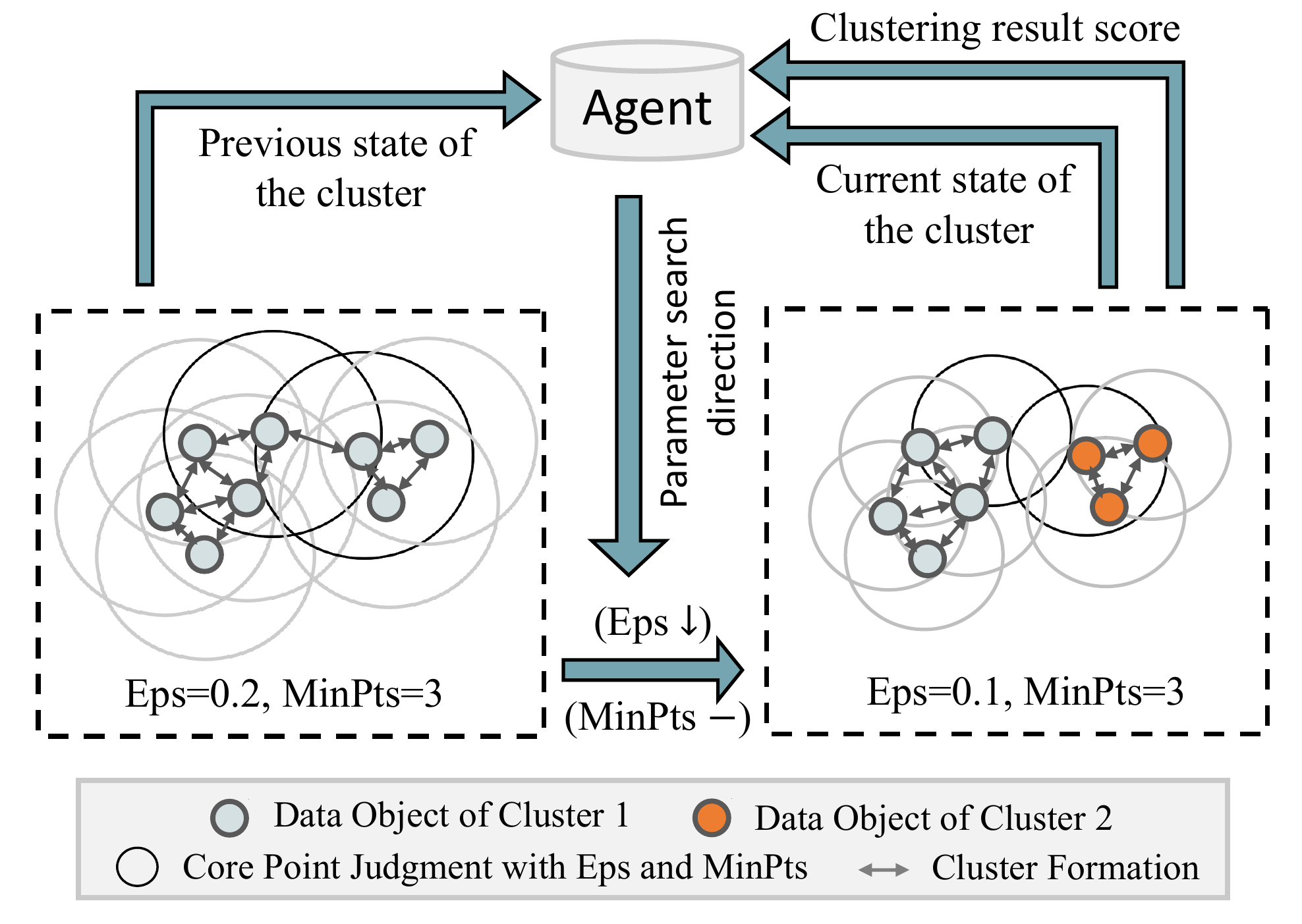}
  \vspace{-1.2mm}
  \caption{Markov process of parameter search. The agent uses the data as the environment, and determines the action to search by observing the clustering state and reward.}\label{fig:intro}
  \vspace{-3.2mm}
\end{figure}

However, the two global parameters of DBSCAN, the distance of the cluster formation $Eps$ and the minimum data objects required inside the cluster $MinPts$, that need to be manually specified, bring \textbf{Three} challenges to its clustering process.
\textbf{First, parameters free challenge.}
$Eps$ and $MinPts$ have considerable influence on the clustering effect, but it needs to be determined a priori.
The method based on the $k$-distance~\cite{lu2007vdbscan,Mitra2011kddclus} estimates the possible values of the $Eps$ through significant changes in the curve, but it still needs to manually formulate $Minpts$ parameters in advance.
Although some improved DBSCAN methods avoid the simultaneous adjustment of $Eps$ and $MinPts$ to varying degrees, but they also necessary to pre-determine the cutoff distance parameter~\cite{diao2018lpdbscsan}, grid parameter~\cite{darong2012grid}, Gaussian distribution parameter~\cite{smiti2012dbscangm} or fixed $Minpts$ parameter~\cite{hou2016dsets,akbari2016outlier}.
Therefore, the first challenge is how to perform DBSCAN clustering without tuning parameters based on expert knowledge.
\textbf{Second, adaptive policy challenge.}
Due to the different data distributions and cluster characteristics in clustering tasks, traditional DBSCAN parameter searching methods based on fixed patterns~\cite{lu2007vdbscan, Mitra2011kddclus} encounter bottlenecks in the face of unconventional data problems.
Moreover, hyperparameter optimization methods~\cite{karami2014bdedbscan, bergstra2011tpe, lessmann2005ga} using external clustering evaluation index based on label information as objective functions are not effective in the absence of data label information.
The methods~\cite{lai2019mvodbscan,zhou2014silhouette_coefficient} that only use the internal clustering evaluation index as the objective function, are limited by the accuracy problem, despite that they do not require label information.
In addition, for streaming data that needs to be clustered continuously but the data distribution is constantly changing, the existing DBSCAN parametric search methods do not focus on how to use past experience to adaptively formulate search policies for newly arrived data.
Thus, how to effectively and adaptively adjust the parameter search policy of the data and be compatible with the lack of label information is regarded as the second challenge.
\textbf{Third, computational complexity challenge.}
Furthermore, the parameter search is limited by the parameter space that cannot be estimated.
Searching too many invalid parameters will increase the search cost~\cite{Anssi2020large_continuous}, and too large search space will bring noise interfering with clustering accuracy~\cite{dulacarnold2016large_discrete}.
Hence, how to quickly search for the optimal parameters while securing clustering accuracy is the third challenge that needs to be addressed.

\tobeupdated{In recent years, Deep Reinforcement Learning (DRL) \cite{scott2018td3,lillicrap2015ddpg} has been widely used for tasks lacking training data due to its ability to learn by receiving feedback from the environment \cite{peng2022reinforced}}.
In this paper, to handle the problem of missing optimal DBSCAN parameter labeling and the challenges above, we propose \textbf{\framework}, a novel, adaptive, recursive \textbf{D}eep \textbf{R}einforcement \textbf{L}earning \textbf{DBSCAN} parameter search framework, to obtain optimal parameters in multiple scenarios and tasks stably.
We first take the cluster changes after each step of clustering as the observable state, the parameter adjustment direction as the action, and transform the parameter search process into a Markov decision process in which the DRL agents autonomously perceive the environment to make decisions (Fig. \ref{fig:intro}).
Then, through weak supervision, we construct reward based on a small number of external clustering indices, and fuse the global state and the local states of multiple clusters based on the attention mechanism, to prompt the agents to learn how to adaptively perform the parameter search for different data.
In addition, to improve the learning efficiency of the policy network, we optimize the base framework through a recursive mechanism based on agents with different search precisions to achieve the highest clustering accuracy of the parameters stably and controllable.
Finally, considering the existence of DBSCAN clustering scenarios with no labels, few labels, initial data, and incremental data, we designed four working modes: retraining mode, continuous training mode, pre-training test mode, and maintenance test mode for compatibility.
We extensively evaluate the parameter search performance of {\framework} with four modes for the offline and online tasks on the public Pathbased, Compound, Aggregation, D31 and Sensor datasets.
The results show that {\framework} can break away from manually defined parameters, automatically and efficiently discover suitable DBSCAN clustering parameters, and has stability in multiple downstream tasks.


The contributions of this paper are summarized as follows:
\textbf{(1)} The first DBSCAN parameter search framework guided by DRL is proposed to automatically select the parameter search directions.
\textbf{(2)} A weakly-supervised reward mechanism and a local cluster state attention mechanism are established to encourage DRL agents to adaptively formulate optimal parameter search policy based on historical experience in the absence of annotations/labels to adapt the data distribution fluctuations.
\textbf{(3)} A recursive DRL parameter search mechanism is designed to provide a fast and stable solution for large-scale parameter space problem.
\textbf{(4)} Extensive experiments in both offline and online tasks are conducted to demonstrate that the four modes of {\framework} have advantages in improving DBSCAN accuracy, stability, and efficiency.


%% file: 2-definition.tex
\section{PROBLEM DEFINITION}

In this section, we give the definitions of DBSCAN clustering, parameter search of DBSCAN clustering, and parameter search in data stream clustering.

\begin{define}
\textbf{(DBSCAN clustering).}
The DBSCAN clustering is the process of obtaining clusters $\mathcal{C}=\{c_{1}, ..., c_{n}, c_{n+1}, ...\}$ for all data objects $\{v_{1}, ..., v_{j}, v_{j+1}, ...\}$ in a data block $\mathcal{V}$ based on the parameter combination $\boldsymbol{P}=\{Eps, MinPts\}$.
Here, $Eps$ is the maximum distance that two adjacent objects can form a cluster, and $MinPts$ refers to the minimum number of adjacent objects within $Eps$ that an object can be a core point.
The formation of the clusters can be understood as the process of connecting the core points to its adjacent objects within $Eps$ \cite{ester1996dbscan} (as shown in Fig. \ref{fig:intro}).
\end{define}


\begin{define}
\textbf{(Parameter search in offline DBSCAN clustering).}
Given the data block $\mathcal{V}=\{v_{1}, ..., v_{j}, v_{j+1}, ...\}$, the parameter search of DBSCAN is the process of finding the optimal parameter combination $\boldsymbol{P}=\{Eps, MinPts\}$ for clustering in all possible parameter spaces.
Here, the feature set $\mathcal{X}$ of data objects in block $\mathcal{V}$ is $\{x_{1}, ..., x_{j},$ $x_{j+1}, ...\}$.
\end{define}

\begin{define}
\textbf{(Parameter search in online DBSCAN clustering).}
Given continuous and temporal $T$ data blocks $\{\mathcal{V}_{1}, ..., \mathcal{V}_{t},$ $\mathcal{V}_{t+1}, ...\}$ as the online data stream, we define the parameter search in online clustering as the process of obtaining the parameter combination $\boldsymbol{P}_{t}=\{Eps_{t}, MinPts_{t}\}$ of the data block $\mathcal{V}_t=\{v_{t,1}, ..., v_{t,j},$ $v_{t,j+1}, ...\}$ at each time $t \in T$.
\end{define}

%% file: 3-model.tex
\section{{\framework} FRAMEWORK}

\begin{figure*}[h]
  \centering
  \includegraphics[width=0.99\textwidth]{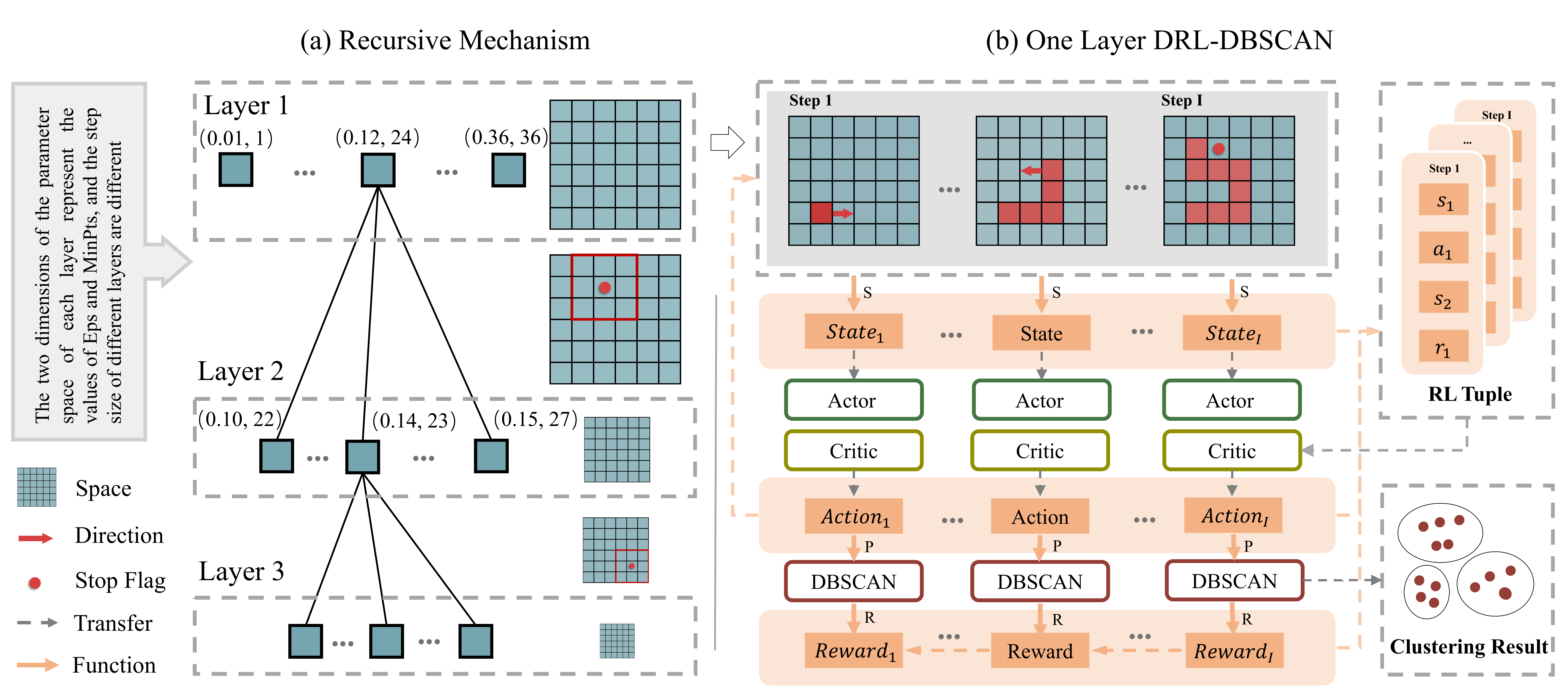}
  \vspace{-1.2mm}
  \caption{{The core model of {\framework}}. (a) Recursive mechanism, takes $3$-layer $6\times6$ parameter space as an example, with layerwise decreasing parameter space. (b) One layer {\framework}, takes the search process in the $1$-th layer of the recursive mechanism as an example, aims to obtain the optimal parameter combination in the parameter space of layer $1$.}    
  \vspace{-1.2mm}
  \label{fig:framework}
\end{figure*}

\begin{figure*}[t]
  \centering
  \includegraphics[width=0.98\textwidth]{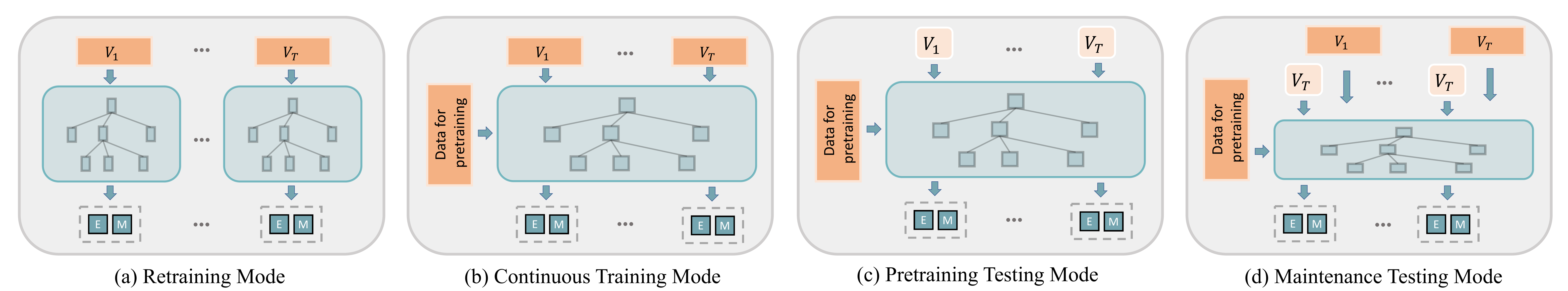}
  \vspace{-1.2mm}
  \caption{{Four working modes. Dark orange square means partially labeled data, light orange square means unlabeled data.}}  \label{fig:task}   
  \vspace{-1.2mm}
\end{figure*}

\tobeupdated{The proposed {\framework} has a core model (Fig.~\ref{fig:framework}) and four working modes (Fig.~\ref{fig:task}) that can be extended to downstream tasks.}
We firstly describe the basic Markov decision process for parameter search (Sec.~\ref{sec:parameter_search}), and the definition of the clustering parameter space and recursion mechanism (Sec.~\ref{sec:recursion}).
Then, we explain the four {\framework} working modes (Sec.~\ref{sec:mode}).


\subsection{Parameter Search with DRL}\label{sec:parameter_search}

Faced with various clustering tasks, the fixed DBSCAN parameter search policy no longer has flexibility.
We propose an automatic parameter search framework {\framework} based on Deep Reinforcement Learning (DRL), in which the core model can be expressed as a Markov Decision Process $MDP(\mathcal{S}, \mathcal{A}, \mathcal{R}, \mathcal{P})$ including state set, action space, reward function and policy optimization algorithm~\cite{mundhenk2000mdp}.
This process transforms the DBSCAN parameter search process into a maze game problem~\cite{zheng2012maze,bom2013pac_man} in the parameter space, aiming to train an agent to search for the end point parameters step by step from the start point parameters by interacting with the environment, and take the end point (parameters in the last step) as the final search result of an episode of the game (as shown in Fig.~\ref{fig:framework}).
Specifically, the agent regards the parameter space and DBSCAN clustering algorithm as the environment, the search position and clustering result as the state, and the parameter adjustment direction as the action.
In addition, a small number of samples are used to reward the agent with exceptional behavior in a weakly supervised manner.
We optimize the policy of agent based on the Actor-Critic \cite{konda2000actor_critic} architecture.
Specifically, the search process for episode $e(e=1,2,...)$ is formulated as follows:

\emph{\textbf{$\bullet$ State:}}
Since the state needs to represent the search environment at each step as accurately and completely as possible, we consider building the representation of the state from two aspects.

Firstly, for the overall searching and clustering situation, we use a 7-tuple to describe the global state of the $i$-th step $(i=1,2,...)$:
\begin{equation}\label{eq:global_state}
\boldsymbol{s}_{global}^{(e)(i)}= \boldsymbol{P}^{(e)(i)} \ {\cup}\ \mathcal{D}_{b}^{(e)(i)} \ {\cup}\  \big\{{R}_{cn}^{(e)(i)}\big\}.
\end{equation}
Here, $ \boldsymbol{P}^{(e)(i)}=\{Eps^{(e)(i)}$, $MinPts^{(e)(i)}\}$ is the current parameter combination.
$\mathcal{D}_{b}^{(e)(i)}$ is the set of quaternary distances, including the distances of $Eps^{(e)(i)}$ from its space boundaries ${B}_{Eps,1}$ and ${B}_{Eps,2}$, the distances of $MinPts^{(e)(i)}$ from its boundaries ${B}_{MinPts,1}$ and ${B}_{MinPts,2}$, 
${R}_{cn}^{(e)(i)}=\frac{|\mathcal{C}^{(e)(i)}|}{|\mathcal{V}|}$ is the ratio of the number of clusters $|\mathcal{C}^{(e)(i)}|$ to the total object number of data block $|\mathcal{V}|$.
Here, the specific boundaries of parameters will be defined in Sec. ~\ref{sec:recursion}.

Secondly, for the description of the situation of each cluster, we define the $\{d+2\}$-tuple of the $i$-th local state of cluster $\boldsymbol{c}_{n} \in \mathcal{\mathcal{C}}$ as:
\begin{equation}\label{eq:local_state}
\boldsymbol{s}_{local, n}^{(e)(i)}= \mathcal{X}_{cent,n}^{(e)(i)} \ {\cup}\  \big\{{D}^{(e)(i)}_{cent,n},\  |\boldsymbol{c}_{n}^{(e)(i)}|\big\}.
\end{equation}
Here, $\mathcal{X}_{cent,n}^{(e)(i)}$ is the central object feature of the $\boldsymbol{c}_{n}$, and $d$ is its feature dimension.
${D}^{(e)(i)}_{cent,n}$ is the Euclidean distance from the cluster center object to the center object of the entire data block.
$|\boldsymbol{c}_{n}^{(e)(i)}|$ means the number of objects contained in cluster $\boldsymbol{c}_{n}$.

Considering the change of the number of clusters at different steps in the parameter search process, we use the Attention Mechanism \cite{vaswani2017attention} to encode the global state and multiple local states into a fixed-length state representation:
\begin{equation}\label{eq:state}
\boldsymbol{s}^{(e)(i)}= 
\sigma \Big(\boldsymbol{F}_{G}(\boldsymbol{s}_{global}^{(e)(i)}) \ {\mathbin\Vert}\  
\sum_{\boldsymbol{c}_n \in \mathcal{C}} \boldsymbol{\alpha}_{att,n} \cdot 
\boldsymbol{F}_{L}(\boldsymbol{s}_{local, n}^{(e)(i)})\Big),
\end{equation}
where $\boldsymbol{F}_{G}$ and $\boldsymbol{F}_{L}$ are the Fully-Connected Network (FCN) for the global state and the local state, respectively.
$\sigma$ represents the \texttt{ReLU} activation function.
And $||$ means the operation of splicing.
$\alpha_{att,n}$ is the attention weight of cluster $\boldsymbol{c}_{n}$, which is formalized as follows:
\begin{equation}
\boldsymbol{\alpha}_{att,n} = \frac{ \sigma \Big(
\boldsymbol{F}_{S}\big(\boldsymbol{F}_{G}(\boldsymbol{s}_{global}^{(e)(i)}) \ {\mathbin\Vert}\  \boldsymbol{F}_{L}(\boldsymbol{s}_{local,n}^{(e)(i)})\big)\Big)}
{\sum_{\boldsymbol{c}_n \in \mathcal{C}} \sigma \Big(\boldsymbol{F}_{S} \big(\boldsymbol{F}_{G}(\boldsymbol{s}_{global}^{(e)(i)}) \ {\mathbin\Vert}\  \boldsymbol{F}_{L}(\boldsymbol{s}_{local,n}^{(e)(i)})\big)\Big)}.
\label{equation:att}
\end{equation}
We concatenate the global state with the local state of each cluster separately, input it into a fully connected network $\boldsymbol{F}_{S}$ for scoring, and use the normalized score of each cluster as its attention coefficient.
This approach establishes the attention to the global search situation when local clusters are expressed.
At the same time, it also makes different types of cluster information have different weights in the final state expression, which increases the influence of important clusters on the state.

\emph{\textbf{$\bullet$ Action:}}
The action $\boldsymbol{a}^{(e)(i)}$ for the $i$-th step is the search direction of parameter.
We define the action space $\mathcal{A}$ as $\{left,right,$ $down,up,stop\}$, where $left$ and $right$ means to reduce or increase the $Eps$ parameter, $down$ and $up$ means to reduce or increase the $MinPts$ parameter, and $stop$ means to stop searching.
Specifically, we build an Actor \cite{konda2000actor_critic} as the policy network to decide action $\boldsymbol{a}^{(e)(i)}$ based on the current state $\boldsymbol{s}^{(e)(i)}$:
\begin{equation}\label{eq:action}
\boldsymbol{a}^{(e)(i)}= Actor(\boldsymbol{s}^{(e)(i)}).
\end{equation}
Here, the $Actor$ is a three-layer Multi-Layer Perceptron (MLP).

In addition, the action-parameter conversion process from the $i$-th step to the $i+1$-th step is defined as follows.
\begin{equation}\label{eq:action2parameter}
\boldsymbol{P}^{(e)(i)} \xrightarrow[]{\boldsymbol{a}^{(e)(i)},\ \boldsymbol{\theta}} \boldsymbol{P}^{(e)(i+1)}.
\end{equation}
Here, $\boldsymbol{P}^{(e)(i)}$ and $\boldsymbol{P}^{(e)(i+1)}$ are the parameter combinations $\{Eps^{(e)(i)},$ $MinPts^{(e)(i)}\}$ and $\{Eps^{(e)(i+1)}, MinPts^{(e)(i+1)}\}$ of the $i$-th step and the $i+1$-th step, respectively.
$\boldsymbol{\theta}$ is the increase or decrease step size of the action.
We discuss the step size in detail in Sec.~\ref{sec:recursion}.
Note that when an action causes a parameter to go out of bounds, the parameter is set to the boundary value and the corresponding boundary distance is set to $-1$ in the next step.

\emph{\textbf{$\bullet$ Reward:}}
Considering that the exact end point parameters are unknown but rewards need to be used to motivate the agent to learn a better parameter search policy, we use a small number of samples of external metrics as the basis for rewards.
We define the immediate reward function of the $i$-th step as:
\begin{equation}\label{eq:reward_function}
\mathcal{R}(\boldsymbol{s}^{(e)(i)},\boldsymbol{a}^{(e)(i)}) = {NMI}\big(DBSCAN(\mathcal{X},\boldsymbol{P}^{(e)(i+1)}),\mathcal{Y}')\big).
\end{equation}
Here, ${NMI}(,)$ stands for the external metric function, Normalized Mutual Information (NMI) \cite{estevez2009nmi} of $DBSCAN$ clustering.
$\mathcal{X}$ is the feature set.
$\mathcal{Y}'$ is a set of partial labels of the data block.
Note that the labels are only used in the training process, and the testing process performs search on unseen data blocks without labels.

In addition, the excellent parameter search action sequence for one episode is to adjust the parameters in the direction of the optimal parameters, and stop the search at the optimal parameters.
Therefore, we consider using both the maximum immediate reward for subsequent steps and the endpoint immediate reward as the reward in the $i$-th step:
\begin{equation}\label{eq:reward}
\boldsymbol{r}^{(e)(i)}=\beta \cdot \max \big\{\mathcal{R}(\boldsymbol{s}^{(e)(m)},\boldsymbol{a}^{(e)(m)})\big\}|_{m=i}^{I}
+\delta \cdot \mathcal{R}(\boldsymbol{s}^{(e)(I)},\boldsymbol{a}^{(e)(I)}), 
\end{equation}
where $\mathcal{R}(\boldsymbol{s}^{(e)(I)},\boldsymbol{a}^{(e)(I)})$ is the immediate rewards for the $I$-th step end point parameters.
And \texttt{max} is used to calculate the future maximum immediate reward before stopping the search in current episode $e$.
$\beta$ and $\delta$ are the impact factors of reward, where $\beta=1-\delta$. 

\emph{\textbf{$\bullet$ Termination:}}
We determine the termination conditions for a complete episode search process as follows:
\begin{equation}\label{eq:termination}
\left\{
\begin{array}{ll}
    \min(\mathcal{D}_{b}^{(e)(i)})<0,      &      \text{Out of bounds stop,}\\
    i>=I_{max},    &     \text{Timeout stop,}\\
    \boldsymbol{a}^{(e)(i)}=stop, where \ i \ge 2,     &    \text{Active stop.}\\
\end{array} 
\right.
\end{equation}
Here, $I_{max}$ is the maximum search step size in an episode.

\emph{\textbf{$\bullet$ Optimization:}}
The parameter search process in the episode $e$ is expressed as: 
$1)$ observe the current state $\boldsymbol{s}^{(e)(i)}$ of DBSCAN clustering;
$2)$ and predict the action $\boldsymbol{a}^{(e)(i)}$ of the parameter adjustment direction based on $\boldsymbol{s}^{(e)(i)}$ through the $Actor$;
$3)$ then, obtain the new state $\boldsymbol{s}^{(e)(i+1)}$;
$4)$ repeat the above process until the end of episode, and get reward $\boldsymbol{r}^{(e)(i)}$ for each step.
The core element of the $i$-th step is extracted as:
\begin{equation}\label{eq:core_element}
\mathcal{T}^{(e)(i)}=(\boldsymbol{s}^{(e)(i)},\boldsymbol{a}^{(e)(i)},\boldsymbol{s}^{(e)(i+1)},\boldsymbol{r}^{(e)(i)}).
\end{equation}
We put $\mathcal{T}$ of each step into the memory buffer and sample $M$ core elements to optimize the policy network $Actor$, and define the key loss functions as follows:
\begin{equation}\label{eq:loss_critic}
    \mathcal{L}_{c} = {\sum}_{\mathcal{T} \in buffer}^{M} \big(\boldsymbol{r}^{(e)(i)}+\gamma \cdot Critic(\boldsymbol{s}^{(e)(i+1)},\boldsymbol{a}^{(e)(i+1)})- \notag
\end{equation}
\begin{equation}
    Critic(\boldsymbol{s}^{(e)(i)},\boldsymbol{a}^{(e)(i)})\big)^{2},
\end{equation}
\begin{equation}\label{eq:loss_actor}
    \mathcal{L}_{a} = -\frac{{\sum}_{\mathcal{T} \in buffer}^{M} Critic\big(\boldsymbol{s}^{(e)(i)},Actor(\boldsymbol{s}^{(e)(i)})\big)}{M}. 
\end{equation}
Here, we define a three-layer MLP as the $Critic$ to learn the action value of state \cite{konda2000actor_critic}, which is used to optimize the $Actor$.
And $\gamma$ means reward decay factor.
Note that we use the policy optimization algorithm named Twin Delayed Deep Deterministic strategy gradient algorithm (TD3)~\cite{scott2018td3} in our framework, and it can be replaced with other DRL policy optimization algorithms \cite{lillicrap2015ddpg, konda2000actor_critic}.

\subsection{Parameter Space and Recursion Mechanism}\label{sec:recursion}

In this section, we will define the parameter space of the agent proposed in the previous section and the recursive search mechanism based on different parameter spaces.
Firstly, in order to deal with the fluctuation of the parameter range caused by different data distributions, we normalize the data features, thereby transforming the maximum $Eps$ parameter search range into the $(0,\sqrt{d}]$ range.
Unlike $Eps$, the $MinPts$ parameter must be an integer greater than $0$.
Therefore, we propose to delineate a coarse-grained $MinPts$ maximum parameter search range according to the size or dimension of the dataset.
Subsequently, considering the large parameter space affects the efficiency when performing high-precision parameter search, we propose to use a recursive mechanism to perform a progressive search.
The recursive process is shown in Fig.~\ref{fig:framework}.
We narrow the search range and increase the search precision layer by layer, and assign a parameter search agent $agent^{(l)}$ defined in Sec.~\ref{sec:parameter_search} to each layer $l$ to search for the optimal parameter combination $\boldsymbol{P}_{o}^{(l)}=\{Eps_{o}^{(l)}, MinPts_{o}^{(l)}\}$ under the requirements of the search precision and range of the corresponding layer.
The minimum search boundary ${B}_{p,1}^{(l)}$ and maximum search boundary ${B}_{p,2}^{(l)}$ of parameter $p \in \{Eps, MinPts\}$ in the $l$-th layer $(l=1,2,...)$ are defined as:
\begin{equation}\label{eq:boundary}
\begin{split}
& {B}_{p,1}^{(l)} : \max \big\{ {B}_{p,1}^{(0)}, \ p_{o}^{(l-1)} - \frac{\pi_{p}}{2} \cdot \theta_{p}^{(l)} \big\},\\
& {B}_{p,2}^{(l)} : \min \big\{ p_{o}^{(l-1)} + \frac{\pi_{p}}{2} \cdot \theta_{p}^{(l)}, {B}_{p,2}^{(0)} \big\}. \\
\end{split}
\end{equation}
Here, $\pi_{p}$ is the number of searchable parameters in the parameter space of parameter $p$ in each layer.
${B}_{p,1}^{(0)}$ and ${B}_{p,2}^{(0)}$ are the space boundaries of parameter $p$ in the $0$-th layer, which define the maximum parameter search range.
$p_{o}^{(l-1)} \in \boldsymbol{P}_{o}^{(l-1)}$ is the optimal parameter searched by the previous layer, and $p_{o}^{(0)} \in \boldsymbol{P}_{o}^{(0)}$ is the midpoint of ${B}_{p,1}^{(0)}$ and ${B}_{p,2}^{(0)}$.
In addition, $\theta_{p}^{(l)}$ is the search step size, that is, the search precision of the parameter $p$ in the $l$-th layer, which is defined as follows:

\begin{equation}\label{eq:theta}
\theta_{p}^{(l)} =
\left\{
\begin{array}{ll}
    \frac{\theta_{p}^{(l-1)}}{\pi_{p}},      & {if \ p \ = \ Eps;}\\
    \max \big\{\lfloor \frac{\theta_{p}^{(l-1)}}{\pi_{p}} + \frac{1}{2} \rfloor, \ 1 \big\},     & {otherwise.}\\
\end{array} 
\right.
\end{equation}
Here, $\theta_{p}^{(l-1)}$ is the step size of the previous layer and $\theta_{p}^{(0)}$ is the size of the parameter maximum search range.
$\lfloor \rfloor$ means round down.

\vspace{0.5mm}
\noindent \emph{\tobeupdated{\textbf{Complexity Discussion:}}}
It is known that the minimum search step size of the recursive structure with layers $L$ is $\theta_{p}^{(L)}$.
Then the computational complexity when there is no recursive structure is $O(N)$, where $N$ the size of the parameter space $\theta_{p}^{(0)} / \theta_{p}^{(L)}=(\pi_{p})^{L}$.
And {\framework} with $L$-layer recursive structure only takes $L \cdot (\pi_{p})$, reducing the complexity from $O(N)$ to $O(log\ N)$.

\subsection{Proposed {\framework}}\label{sec:mode}
Algorithm~\ref{algorithm:train} shows the process of the proposed {\framework} core model.
Given a data block $\mathcal{V}$ with partial label $\mathcal{Y}'$, the training process repeats the parameter search process (Lines~\ref{traincode:observe_state}-\ref{traincode:termination}) for multiple episodes at each layer to optimize the agent (Line~\ref{traincode:loss}).
In this process, we update the optimal parameter combination (Line~\ref{traincode:optimal1} and Line~\ref{traincode:optimal2} based on the immediate reward (Eq. (\ref{eq:reward_function})).
In order to improve efficiency, we build a hash table to record DBSCAN clustering results of searched parameter combinations.
In addition, we established early stopping mechanisms to speed up the training process when the optimal parameter combination does not change (Line~\ref{traincode:stop1} and Line~\ref{traincode:stop2}).
It is worth noting that the testing process uses the trained agents to search directly with one episode, and does not set the early stop.
Furthermore, the testing process does not need labels, and the end point parameters of the unique episode of the last layer are used as the final optimal parameter combination.

In order to better adapt to various task scenarios, we define four working modes of {\framework} as shown in Fig.~\ref{fig:task}. Their corresponding definitions are as follows:
\textbf{(1) Retraining Mode ($DRL_{re}$).}
The optimal parameters are searched based on the training process. When the dataset changes, the agent at each layer is reinitialized.
\textbf{(2) Continuous Training Mode ($DRL_{con}$).}
The agents are pre-trained in advance. When the dataset changes, continue searching based on the training process using the already trained agents.
\textbf{(3) Pretraining Testing Mode ($DRL_{all}$).}
The agents are pre-trained in advance. When the dataset changes, searching directly based on the testing process without labels.
\textbf{(4) Maintenance Testing Mode  ($DRL_{one}$).}
The agents are pre-trained in advance. When the dataset changes, searching directly based on the testing process without labels. After pre-training, regular maintenance training is performed with labeled historical data.


\begin{algorithm}[t]
\SetAlgoRefName{1}
\SetAlgoVlined
\KwIn{The features $\mathcal{X}$ and partial labels $\mathcal{Y}'$ of block $\mathcal{V}$; 
Agents for each layer: $\{agent^{(l)}\}|_{l=1}^{L_{max}}$; 
}
\KwOut{Optimal parameter combination: $\boldsymbol{P}_{o}$;}
\For{$l = 1, ... , L_{max}$} {
    Initialize parameter space via Eq.~(\ref{eq:boundary}) and Eq.~(\ref{eq:theta}); \label{traincode:boundary}\\
    \For{$e = 1, ... , E_{max}$} {
        Initialize $\boldsymbol{P}^{(e)(0)}$ by $\boldsymbol{P}_{o}^{(l-1)}$;\\
        \For{$i = 1, ... , I_{max}$} {
            Obatin the current state $\boldsymbol{s}^{(e)(i)}$ via Eq. (\ref{eq:state}); \label{traincode:observe_state}\\
            Choose the action $\boldsymbol{a}^{(e)(i)}$ via Eq. (\ref{eq:action}); \label{traincode:choose_action}\\
            Get new parameters $\boldsymbol{P}^{(e)(i)}$ via Eq. (\ref{eq:action2parameter}); \label{traincode:get_parameter}\\
            Clustering using the current parameters; \label{traincode:dbsacn_cluster}\\
            Termination judgment via Eq. (\ref{eq:termination}); \label{traincode:termination}\\
        }
        \If{is TRAIN}{
        Get rewards $\boldsymbol{r}^{(e)(i)}$ via Eq. (\ref{eq:reward}), $\forall i \in \{1, I\}$; \label{traincode:reward}\\
        Store $\mathcal{T}^{(e)(i)}$ in buffer \label{traincode:store_t} via Eq. (\ref{eq:core_element}), $\forall i \in \{1, I\}$;\\
        Sampling and learning via Eq. (\ref{eq:loss_actor}) and Eq. (\ref{eq:loss_critic}); \label{traincode:loss}\\
        }
        Update optimal parameter combination $\boldsymbol{P}_{o}^{(l)}$; \label{traincode:optimal1}\\
        Early stop judgment; \label{traincode:stop1}\\
    }
    Update optimal parameter combination $\boldsymbol{P}_{o}$; \label{traincode:optimal2}\\
    Early stop judgment; \label{traincode:stop2}\\
}
\caption{The core model of {\framework}}
\label{algorithm:train}
\end{algorithm}

%% file: 4-experiment.tex
\section{EXPERIMENTS}

In this section, we conduct experiments mainly including the following: \textbf{(1)} performance comparison over {\framework} and baseline in offline tasks, and explanation of the search process for Reinforcement Learning (RL) (Sec. \ref{sec:offline});
\textbf{(2)} performance analysis of {\framework} and its variants, and advantage comparison between four working modes in online tasks (Sec. \ref{sec:online});
\textbf{(3)} sensitivity analysis of hyperparameters and their impact on the model (Sec. \ref{sec:hyperparameter}).

\subsection{Experiment Setup}\label{sec:setup}

\vspace{0.5mm}
\noindent \emph{\textbf{Datasets}}.\label{sec:dataset}
To fully analyze our framework, the experimental dataset consists of $4$ artificial clustering benchmark datasets and $1$ public real-world streaming dataset (Table \ref{tab:dataset}).
The benchmark datasets \cite{Pasi2018benchmark} are the $2D$ shape sets, including: \textbf{Aggregation} \cite{gionis2007aggregation}, \textbf{Compound} \cite{zahn1971compound}, \textbf{Pathbased} \cite{chang2008pathbased}, and \textbf{D31} \cite{veenman2002d31}.
They involve multiple density types such as clusters within clusters, multi-density, multi-shape, closed loops, etc., and have various data scales.
Furthermore, the real-world streaming dataset \textbf{Sensor} \cite{zhu2010stream_dataset} comes from consecutive information (temperature, humidity, light, and sensor voltage) collected from $54$ sensors deployed by the Intel Berkeley Research Lab.
We use a subset of $80, 864$ for experiments and divide these objects into $16$ data blocks ($\mathcal{V}_1, ..., \mathcal{V}_{16}$) as an online dataset.

\vspace{0.5mm}
\noindent \emph{\textbf{Baseline and Variants}}.\label{sec:baseline_variants}
We compare proposed {\framework} with three types of baselines: (1) traditional hyperparameter search schemes: random search algorithm \textbf{Rand} \cite{bergstra2012rand}, Bayesian optimization based on Tree-structured Parzen estimator algorithm \textbf{BO-TPE} \cite{bergstra2011tpe}; (2) meta-heuristic optimization algorithms: the simulated annealing optimization \textbf{Anneal} \cite{kirkpatrick1983anneal}, particle swarm optimization \textbf{PSO} \cite{shi1998pso}, genetic algorithm \textbf{GA} \cite{lessmann2005ga}, and differential evolution algorithm \textbf{DE} \cite{qin2008de}; (3) existing DBSCAN parameter search methods: \textbf{KDist} (V-DBSCAN) \cite{lu2007vdbscan} and \textbf{BDE}-DBSCAN \cite{karami2014bdedbscan}.
The detailed introduction to the above methods are given in Sec. \ref{sec:related_work}.
We also implement four variants of {\framework} to analysis of state, reward and recursive mechanism settings in Sec. \ref{sec:parameter_search}.
Compared with {\framework}, $DRL_{no-att}$ does not join local state based on the attention mechanism, $DRL_{only-max}$ only uses the maximum future immediate reward as final reward, $DRL_{recu}$ has no early stop mechanism, and $DRL_{recu}$ has no recursion mechanism.




\vspace{0.5mm}
\noindent \emph{\textbf{Implementation Details}}.
For all baselines, we use open-source implementations from the benchmark library Hyperopt \cite{bergstra2013hyperopt} and Scikit-opt \cite{_2022scikitopt}, or provided by the author.
All experiments are conducted on Python 3.7, $36$ core $3.00$GHz Intel Core $i9$ CPU, and NVIDIA RTX $A6000$ GPUs.

\vspace{0.5mm}
\noindent \emph{\textbf{Experimental Setting}}.
The evaluation of {\framework} is based on the four working modes proposed in Sec. ~\ref{sec:mode}.
Considering the randomness of most algorithms, all experimental results we report are the means or variances of $10$ runs with different seeds (except KDist because it's heuristic and doesn't involve random problems).
Specifically, for the pre-training and maintenance training processes of {\framework}, we set the maximum number of episodes $E_{max}$ to $50$, and do not set the early stop mechanism.
For the training process for searching, we set the maximum number of episodes $E_{max}$ to $15$.
In offline tasks and online tasks, the maximum number of recursive layers $L_{max}$ is $3$ and $6$, respectively, and the maximum search boundary in the $0$-th layer of $MinPts$ ${B}_{MinPts,2}^{(0)}$ is $0.25$ and $0.0025$ times the size of block, respectively.
In addition, we use the unified label training proportion $0.2$, the $Eps$ parameter space size $\pi_{Eps}$ $5$, the $MinPts$ parameter space size $\pi_{MinPts}$ $4$, the maximum number of search steps $I_{max}$ $30$ and the reward factor $\delta$ $0.2$. 
\tobeupdated{The FCN and MLP dimensions are uniformly set to $32$ and $256$, the reward decay factor $\gamma$ of Critic is $0.1$, and the number of samples $M$ from the buffer is $16$.}
Furthermore, all baselines use the same objective function (Eq. (\ref{eq:reward_function})), parameter search space, and parameter minimum step size as {\framework} if they support the settings.

\vspace{0.5mm}
\noindent \emph{\textbf{Evaluation Metrics}}.
We evaluate the experiments in terms of accuracy and efficiency.
Specifically, we measure the clustering accuracy based on normalized mutual information (NMI) \cite{estevez2009nmi} and adjusted rand index (ARI) \cite{vinh2010ari}.
For the efficiency, we use the consumed DBSCAN clustering rounds as the measurement.

\subsection{Offline Evaluation}\label{sec:offline}

Offline evaluation is based on four artificial benchmark datasets.
Since there is no data for pre-training in offline scenarios, we only compare the parameter search performance of {\framework} using the retraining mode $DRL_{re}$ with baselines.

\vspace{0.5mm}
\noindent \emph{\textbf{Accuracy and Stability Analysis. }}
In Table \ref{tab:offline_all}, we summarize the means and variances of the NMI and ARI corresponding to the optimal DBSCAN parameter combinations that can be searched by $DRL_{re}$ and baselines within $30$ clustering rounds.
It can be seen from the mean accuracy results of ten runs that in the Pathbased, Compound, Aggregation, and D31 datasets, $DRL_{re}$ can effectively improve the performance of $4\%$ \& $6\%$, $3\%$ \& $3\%$, $20\%$ \& $26\%$ and $5\%$ \& $26\%$ on NMI and ARI, relative to the best performing baselines.
At the same time, as the dataset size increases, the advantage of $DRL_{re}$ compared to other baselines in accuracy gradually increases.
Furthermore, the experimental variances shows that $DRL_{re}$ improves stability by $4\%$ \& $6\%$, $1\%$ \& $1\%$, $9\%$ \& $13\%$ and $2\%$ \& $17\%$ on NMI and ARI, relative to the best performing baselines.
The obvious advantages in terms of accuracy and stability indicate that $DRL_{re}$ can stably find excellent parameter combinations in multiple rounds of parameter search, compared with other hyperparameter optimization baselines under the same objective function.
Besides, $DRL_{re}$ is not affected by the size of the dataset.
Among all the baselines, PSO and DE are relatively worse in terms of accuracy, because their search in the parameter space is biased towards continuous parameters, requiring more rounds to achieve optimal results.
BO-TPE learns previously searched parameter combinations through a probabilistic surrogate model and strikes a balance between exploration and exploitation, with significant advantages over other baselines.
The proposed {\framework} not only narrows the search space of parameters of each layer progressively through a recursive structure, but also learns historical experience, which is more suitable for searching DBSCAN clustering parameter combinations.

\begin{table}[t]
    \setlength{\abovecaptionskip}{0.cm}
    \setlength{\belowcaptionskip}{-0.cm}
    \caption{Characteristics of Datasets.}\label{tab:dataset}
    \centering
    \scalebox{1.0}{
        \begin{tabular}{c|c|ccc|cc}
        \hline
        Type & Dataset & Classes & Size & Dim. & Time\\
        \hline
        \multirow{4}*{Offline} & Pathbased & $3$ & $300$ & $2$ & $\times$\\
        & Compound & $6$ & $399$ & $2$ & $\times$\\
        & Aggregation & $7$ & $788$ & $2$ & $\times$\\
        & D31 & $31$ & $3100$ & $2$ & $\times$\\
        \hline
        Online & Sensor & $54$ & $80640$ & $5$ & $\checkmark$\\
        \hline
        \end{tabular}
    }
\vspace{-4.5mm}
\end{table}

\begin{figure}[t]
\centering
\includegraphics[width=8.5cm]{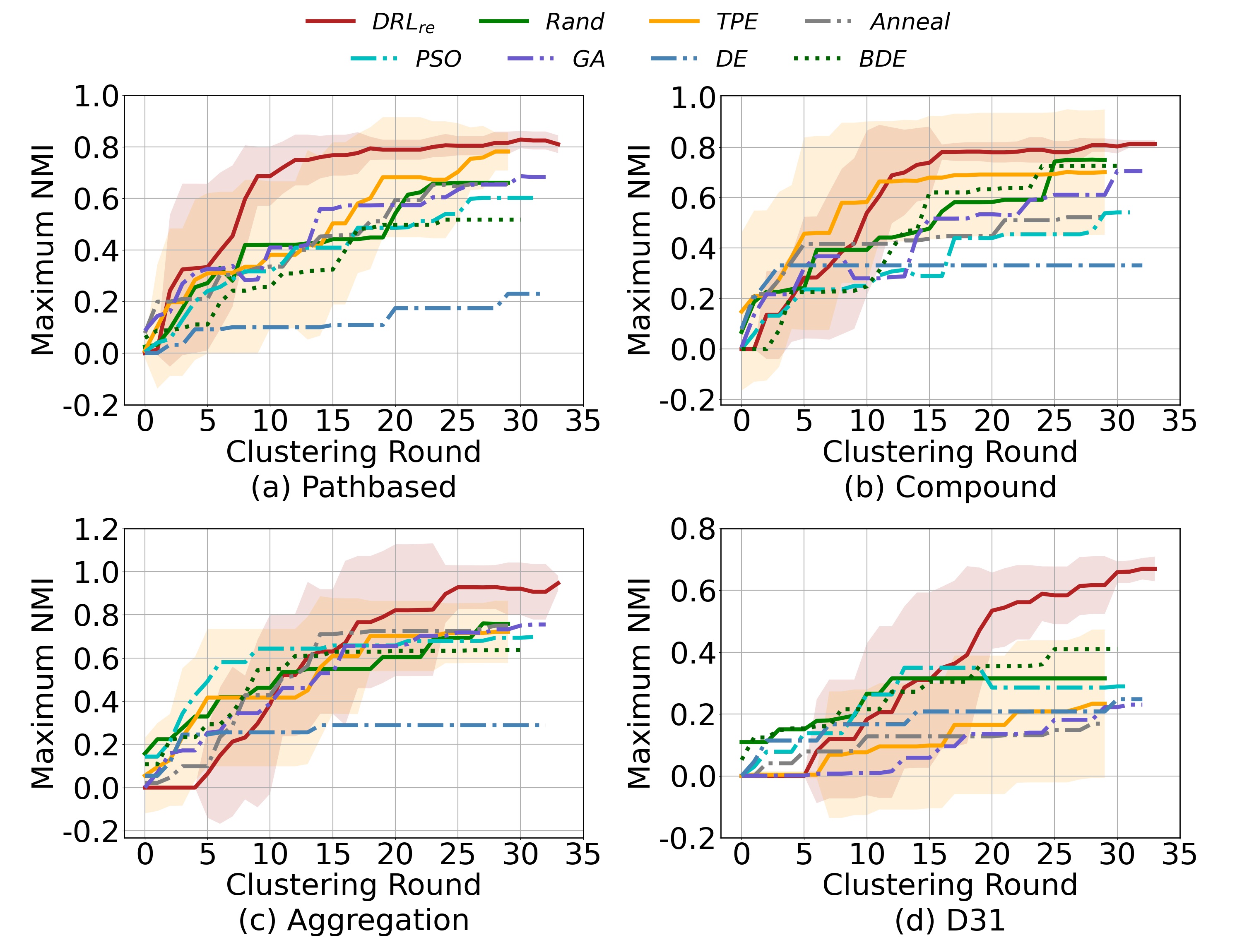}\vspace{-1em}
\centering
\caption{Offline clustering efficiency comparison.}\label{fig:offline_efficiency}
\vspace{-1.2mm}
\end{figure}

\vspace{0.5mm}
\noindent \emph{\textbf{Efficiency Analysis. }}
We present the average historical maximum NMI results for $DRL_{re}$ and all baselines when consuming a different number of clustering rounds in Fig. \ref{fig:offline_efficiency}.
The shade in the figure represents the fluctuation range (variance) of NMI in multiple runs (only BO-TPE is also shown with shade as the representative in the baselines).
The results suggest that in the four datasets, $DRL_{re}$ can maintain a higher speed of finding better parameters than baselines, and fully surpass all baselines after the $5$-th, $12$-th, $17$-th, and $16$-th rounds, respectively.
In the Pathbased dataset, the clustering rounds of $DRL_{re}$ is $2.49$ times faster than that of BO-TPE when the NMI of the parameter combination searched by $DRL_{re}$ reaches $0.72$.
Besides, the results show that, with the increase of clustering rounds, the shadow area of the $DRL_{re}$ curve gradually decreases, while the shadow range of BO-TPE tends to be constant.
The above observation also demonstrates the good stability of the search when the number of rounds of {\framework} reaches a specific number.

\begin{figure}[t]
\centering
\includegraphics[width=8.5cm]{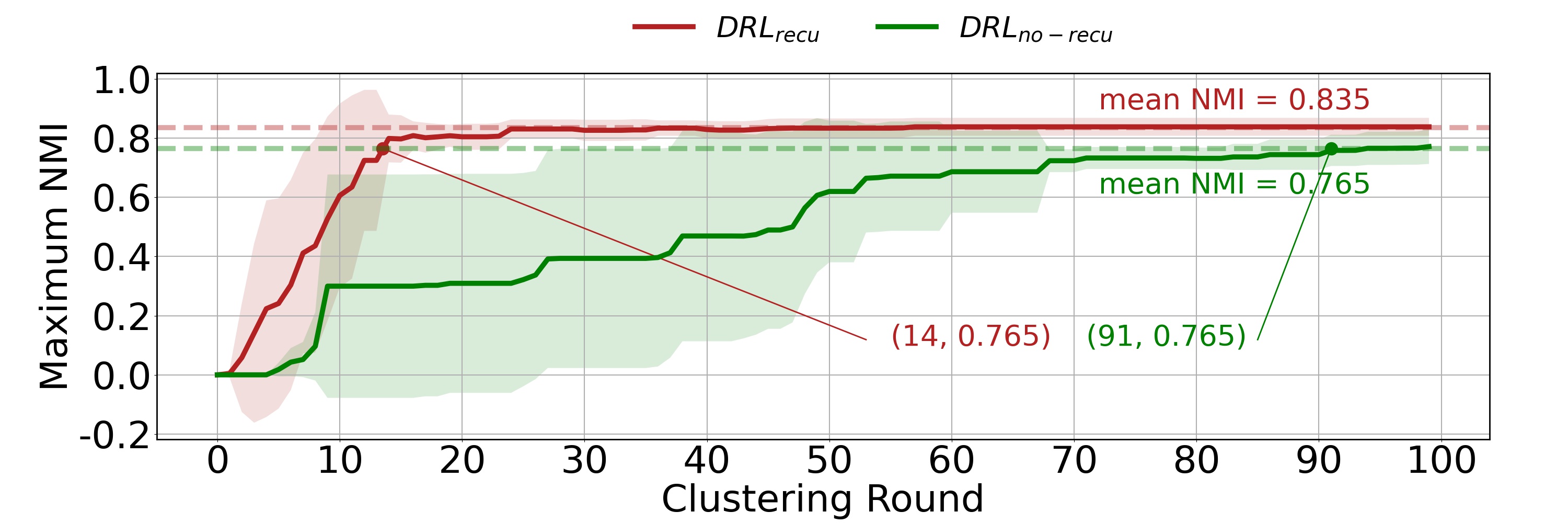}\vspace{-1em}
\centering
\caption{Efficiency comparison of recursive mechanism.}\label{fig:recur_eff}
\vspace{-3.2mm}
\end{figure}

\vspace{0.5mm}
\noindent \emph{\textbf{DRL-DBSCAN Variants.}}
We compare the $DRL_{recu}$ with $DRL_{no-recu}$ which without the recursion mechanism in Fig. \ref{fig:recur_eff}.
Note that, for $DRL_{recu}$, we turn off the early stop mechanism so that it can search longer to better compare with $DRL_{no-recu}$.
The results show that the first $100$ episodes of the recursive mechanism bring the maximum search speedup ratio of $6.5$, which effectively proves the contribution of the recursive mechanism in terms of efficiency.

\begin{table*}[t]
    \setlength{\abovecaptionskip}{0.cm}
    \setlength{\belowcaptionskip}{-0.cm}
    \caption{Offline evaluation performance. \textmd{The best results are bolded and second-best are underlined.}}\label{tab:offline_all}
    \centering
    \scalebox{1.0}{
        \begin{tabular}{c|c|cc|cccc|cc|ccc}
        \hline
        \multirow{2}*{\textbf{Dataset}} & \multirow{2}*{\textbf{Metrics}} & \multicolumn{2}{c|}{\textbf{Traditional}} & \multicolumn{4}{c|}{\textbf{Meta-heuristic}} & \multicolumn{5}{c}{\textbf{Dedicated}} \\
        \cline{3-13}
         &  & \multirow{1}*{\textbf{Rand}} & \multirow{1}*{\textbf{BO-TPE}} & \multirow{1}*{\textbf{Anneal}} & \multirow{1}*{\textbf{PSO}} & \multirow{1}*{\textbf{GA}} & \multirow{1}*{\textbf{DE}} & \multirow{1}*{\textbf{KDist}} & \multirow{1}*{\textbf{BDE}} & \multicolumn{1}{c}{\textbf{DRL$_{re}$}} & (Mean) & (Var.)\\
        \hline
        \hline 
        \rowcolor{gray!15}                             & NMI & .66$\pm$.23 & .\underline{78$\pm$.07} & .65$\pm$.24 & .60$\pm$.28 & .68$\pm$.19 & .22$\pm$.28 & .40$\pm$.- - & .51$\pm$.33 & \textbf{.82$\pm$.03} & $\uparrow$ .04 & $\downarrow$ .04\\
        \rowcolor{gray!15} \multirow{-2}*{Pathbased}   & ARI & .63$\pm$.21 & \underline{.79$\pm$.10} & .66$\pm$.25 & .55$\pm$.38 & .67$\pm$.26 & .18$\pm$.28 & .38$\pm$.- - & .48$\pm$.40 & \textbf{.85$\pm$.04} & $\uparrow$ .06  & $\downarrow$ .06\\
        \rowcolor{white!15}                            & NMI & \underline{.75$\pm$.05} & .70$\pm$.24 & .52$\pm$.36 & .46$\pm$.34 & .70$\pm$.25 & .33$\pm$.35 & .39$\pm$.- - & .72$\pm$.25 & \textbf{.78$\pm$.04} & $\uparrow$ .03  & $\downarrow$ .01\\
        \rowcolor{white!15} \multirow{-2}*{Compound}   & ARI & \underline{.73$\pm$.04} & .68$\pm$.24 & .51$\pm$.35 & .42$\pm$.36 & .68$\pm$.24 & .31$\pm$.34 & .39$\pm$.- - & .70$\pm$.25 & \textbf{.76$\pm$.03} & $\uparrow$ .03  & $\downarrow$ .01\\
        \rowcolor{gray!15}                             & NMI & \underline{.76$\pm$.11} & .72$\pm$.14 & .75$\pm$.27 & .59$\pm$.35 & .75$\pm$.15 & .28$\pm$.37 & .60$\pm$.- - & .63$\pm$.28 & \textbf{.96$\pm$.02} & $\uparrow$ .20 & $\downarrow$ .09\\
        \rowcolor{gray!15} \multirow{-2}*{Aggregation} & ARI & .68$\pm$.16 & .63$\pm$.19 & .\underline{70$\pm$.27} & .51$\pm$.37 & .68$\pm$.19 & .25$\pm$.35 & .52$\pm$.- - & .54$\pm$.28 & \textbf{.96$\pm$.03} & $\uparrow$ .26 & $\downarrow$ .13\\
        \rowcolor{white!15}                            & NMI & .31$\pm$.33 & .23$\pm$.24 & .17$\pm$.19 & .36$\pm$.33 & .23$\pm$.20 & .24$\pm$.26 & .07$\pm$.- - & \underline{.41$\pm$.36} & \textbf{.67$\pm$.02} & $\uparrow$ .26 & $\downarrow$ .17\\
        \rowcolor{white!15} \multirow{-2}*{D31}        & ARI & .14$\pm$.26 & .04$\pm$.05 & .03$\pm$.04 & .09$\pm$.22 & .04$\pm$.04 & .06$\pm$.09 & .00$\pm$.- - & \underline{.21$\pm$.28} & \textbf{.26$\pm$.02} & $\uparrow$ .05 & $\downarrow$ .02\\
        \hline
        \hline
        \end{tabular}
    }
\end{table*}

\vspace{0.5mm}
\noindent \emph{\textbf{Label Proportion Comparison. }}
Considering the influence of the proportion of labels participating in training on the {\framework} training process's reward and the objective function of other baselines, we conduct experiments with different label proportions in Pathbased, and the results are shown in Fig. \ref{fig:mix}(b) (the vertical lines above the histograms are the result variances).
It can be seen that under different label proportions, the average NMI scores of $DRL_{re}$ are better than baselines, and the variance is also smaller.
Additionally, as the proportion of labels decreases, the NMI scores of most of the baselines drop sharply, while the $DRL_{re}$ tends to be more or less constant.
These stable performance results demonstrate the adaptability of {\framework} to label proportion changes.

\vspace{0.5mm}
\noindent \emph{\textbf{Case Study. }}\label{sec:case}
To better illustrate the parameter search process based on RL, we take $3$ episodes in the $3$-rd recursive layer of the Pathbased dataset as an example for case study (Table \ref{tab:case}).
The columns in Table \ref{tab:case} are the action sequences made by the agent $agent^{(l)}$ in different episodes, the termination types of the episodes, the parameter combinations and NMI scores of the end point.
We can observe that {\framework} aims to obtain the optimal path from the initial parameter combination to the optimal parameter combination.
The path-based form of search can use the optimal path learned in the past while retaining the ability to explore the unknown search direction for each parameter combination along the path.
Since we add the $stop$ action to the action, {\framework} can also learn how to stop at the optimal position, which helps extend {\framework} to online clustering situations where there is no label at all.
Note that we save the clustering information of the parameter combinations experienced in the past into a hash table, thereby no additional operations are required for repeated paths.

\begin{table}[t]
    \setlength{\abovecaptionskip}{0.cm}
    \setlength{\belowcaptionskip}{-0.cm}
    \caption{Case study.}\label{tab:case}
    \centering
    \scalebox{1.0}{
        \begin{tabular}{c|c|c|c}
        \hline
        Action Sequence & Stop Type & $Eps$ / $MinPts$ & NMI\\
        \hline
        $stop \rightarrow left \rightarrow up \rightarrow$  & \multirow{3}*{Out} & \multirow{3}*{$0.155$ / $41$} & \multirow{3}*{.64}\\
        $down \rightarrow down \rightarrow right \rightarrow $  &  &  & \\
        $right \rightarrow right$  &  &  & \\
        \hline
        $down \rightarrow right \rightarrow up \rightarrow$  & \multirow{2}*{Active} & \multirow{2}*{$0.139$ / $41$} & \multirow{2}*{.74}\\
        $left \rightarrow down \rightarrow stop$ & & & \\
        \hline
        $left \rightarrow left$  & Out & $0.123$ / $43$ & .81\\
        \hline
        \end{tabular}
    }
\end{table}

\subsection{Online Evaluation}\label{sec:online}

The learnability of RL enables {\framework} to better utilize past experience in online tasks.
To this end, we comprehensively evaluate four working modes of {\framework} on a streaming dataset, Sensor.
Specifically, the first eight blocks of the Sensor are used for the pre-training of $DRL_{con}$, $DRL_{all}$ and $DRL_{one}$, and the last eight blocks are used to compare the results with baselines.
Since baselines cannot perform incremental learning for online tasks, we initialize the algorithm before each block starts like $DRL_{re}$.
In addition, both experiments of $DRL_{all}$ and $DRL_{one}$ use unseen data for unlabeled testing, whereas $DRL_{one}$ uses the labeled historical data for model maintenance training after each block ends testing.

\begin{table*}[t]
    \setlength{\abovecaptionskip}{0.cm}
    \setlength{\belowcaptionskip}{-0.cm}
    \caption{Online evaluation NMI for training-based modes. \textmd{The best results are bolded and second-best are underlined.}}\label{tab:online_train}
    \centering
    \scalebox{1.0}{
        \begin{tabular}{c|cc|cccc|cc|cccc}
        \hline
        Blocks & \multirow{1}*{\textbf{Rand}} & \multirow{1}*{\textbf{BO-TPE}} & \multirow{1}*{\textbf{Anneal}} & \multirow{1}*{\textbf{PSO}} & \multirow{1}*{\textbf{GA}} & \multirow{1}*{\textbf{DE}} & \multirow{1}*{\textbf{KDist}} & \multirow{1}*{\textbf{BDE}} & \multirow{1}*{\textbf{DRL$_{re}$}} & \multirow{1}*{\textbf{DRL$_{con}$}}  & (Mean) & (Var.)\\
        \hline
        \hline 
        \rowcolor{white!15} $\mathcal{V}_{9}$  & .67$\pm$.24 & .83$\pm$.03 & .53$\pm$.37 & .74$\pm$.10 & .65$\pm$.29 & .19$\pm$.31 & .30$\pm$.- - & .70$\pm$.21 & \underline{.86$\pm$.01} & \textbf{.87$\pm$.00}  & $\uparrow$ .04 & $\downarrow$ .03\\
        \rowcolor{white!15} $\mathcal{V}_{10}$ & .36$\pm$.15 & \underline{.50$\pm$.07} & .45$\pm$.17 & \underline{.50$\pm$.20} & .43$\pm$.15 & .15$\pm$.17 & .20$\pm$.- - & .37$\pm$.20 & \underline{.50$\pm$.27} & \textbf{.64$\pm$.06}  & $\uparrow$ .14 & $\downarrow$ .01\\
        \rowcolor{white!15} $\mathcal{V}_{11}$ & .40$\pm$.06 & .43$\pm$.10 & .32$\pm$.26 & .55$\pm$.16 & .43$\pm$.08 & .09$\pm$.12 & .12$\pm$.- - & .47$\pm$.16 & \underline{.60$\pm$.16} & \textbf{.68$\pm$.02}  & $\uparrow$ .13 & $\downarrow$ .04\\
        \rowcolor{white!15} $\mathcal{V}_{12}$ & .44$\pm$.23 & .62$\pm$.16 & .27$\pm$.35 & .66$\pm$.07 & .50$\pm$.24 & .19$\pm$.28 & .11$\pm$.- - & .41$\pm$.31 & \textbf{.75$\pm$.01} & \underline{.72$\pm$.10}  & $\uparrow$ .09 & $\downarrow$ .06\\
        \rowcolor{white!15} $\mathcal{V}_{13}$ & .84$\pm$.06 & .87$\pm$.04 & .72$\pm$.38 & .68$\pm$.26 & .76$\pm$.17 & .38$\pm$.38 & .62$\pm$.- - & .68$\pm$.23 & \textbf{.92$\pm$.02} & \textbf{.92$\pm$.02}  & $\uparrow$ .08 & $\downarrow$ .02\\
        \rowcolor{white!15} $\mathcal{V}_{14}$ & .74$\pm$.12 & \underline{.82$\pm$.04} & .54$\pm$.37 & .63$\pm$.24 & .54$\pm$.24 & .25$\pm$.25 & .55$\pm$.- - & .56$\pm$.25 & .76$\pm$.25 & \textbf{.85$\pm$.00}  & $\uparrow$ .03 & $\downarrow$ .04\\
        \rowcolor{white!15} $\mathcal{V}_{15}$ & .68$\pm$.24 & .76$\pm$.04 & .66$\pm$.34 & .55$\pm$.25 & .62$\pm$.27 & .28$\pm$.32 & .36$\pm$.- - & .72$\pm$.14 & \textbf{.85$\pm$.07} & \underline{.83$\pm$.13}  & $\uparrow$ .17 & -\\
        \rowcolor{white!15} $\mathcal{V}_{16}$ & .73$\pm$.13 & .77$\pm$.09 & .77$\pm$.10 & .40$\pm$.35 & .67$\pm$.22 & .49$\pm$.31 & .11$\pm$.- - & .67$\pm$.19 & \textbf{.86$\pm$.01} & \textbf{.86$\pm$.00}  & $\uparrow$ .09 & $\downarrow$ .09\\
        \hline
        \hline
        \end{tabular}
    }
\end{table*}
\begin{table*}[t]
    \setlength{\abovecaptionskip}{0.cm}
    \setlength{\belowcaptionskip}{-0.cm}
    \caption{Online evaluation NMI for testing-based modes. \textmd{The best results are bolded and second-best are underlined.}}\label{tab:online_test}
    \centering
    \scalebox{1.0}{
        \begin{tabular}{c|cc|cccc|cc|cccc}
        \hline
        Blocks & \multirow{1}*{\textbf{Rand}} & \multirow{1}*{\textbf{BO-TPE}} & \multirow{1}*{\textbf{Anneal}} & \multirow{1}*{\textbf{PSO}} & \multirow{1}*{\textbf{GA}} & \multirow{1}*{\textbf{DE}} & \multirow{1}*{\textbf{KDist}} & \multirow{1}*{\textbf{BDE}} & \multirow{1}*{\textbf{DRL$_{all}$}} & \multirow{1}*{\textbf{DRL$_{one}$}}  & (Mean) & (Var.) \\
        \hline
        \hline 
        \rowcolor{white!15} $\mathcal{V}_{9}$  & .34$\pm$.31 & .49$\pm$.33 & .22$\pm$.34 & .14$\pm$.29 & .27$\pm$.37 & .10$\pm$.26 & .30$\pm$.- - & .54$\pm$.36 & \textbf{.68$\pm$.30} & \textbf{.68$\pm$.30}  & $\uparrow$ .19 & -\\
        \rowcolor{white!15} $\mathcal{V}_{10}$ & .11$\pm$.14 & .28$\pm$.17 & .17$\pm$.21 & .24$\pm$.01 & .20$\pm$.21 & .12$\pm$.18 & .20$\pm$.- - & .28$\pm$.24 & \textbf{.33$\pm$.16} & \textbf{.33$\pm$.15}  & $\uparrow$ .05 & -\\
        \rowcolor{white!15} $\mathcal{V}_{11}$ & .16$\pm$.15 & .29$\pm$.24 & .23$\pm$.18 & \textbf{.33$\pm$.29} & .23$\pm$.23 & .02$\pm$.05 & .12$\pm$.- - & .21$\pm$.22 & .30$\pm$.13 & \underline{.32$\pm$.08}  & - & -\\
        \rowcolor{white!15} $\mathcal{V}_{12}$ & .23$\pm$.25 & .19$\pm$.24 & .10$\pm$.22 & \underline{.38$\pm$.26} & .34$\pm$.27 & .03$\pm$.06 & .11$\pm$.- - & .29$\pm$.27 & \underline{.38$\pm$.17} & \textbf{.46$\pm$.09}  & $\uparrow$ .08 & -\\
        \rowcolor{white!15} $\mathcal{V}_{13}$ & .58$\pm$.35 & \underline{.70$\pm$.24} & .47$\pm$.40 & .44$\pm$.31 & .36$\pm$.28 & .08$\pm$.14 & .62$\pm$.- - & .32$\pm$.26 & .68$\pm$.34 & \textbf{.70$\pm$.27}  & - & -\\
        \rowcolor{white!15} $\mathcal{V}_{14}$ & .36$\pm$.19 & .34$\pm$.28 & .47$\pm$.35 & .37$\pm$.33 & .27$\pm$.25 & .11$\pm$.24 & .55$\pm$.- - & .43$\pm$.28 & \underline{.60$\pm$.27} & \textbf{.62$\pm$.16}  & $\uparrow$ .15 & $\downarrow$ .03\\
        \rowcolor{white!15} $\mathcal{V}_{15}$ & .45$\pm$.35 & .38$\pm$.36 & .37$\pm$.33 & .30$\pm$.34 & .36$\pm$.32 & .09$\pm$.18 & .36$\pm$.- - & .42$\pm$.31 & \underline{.64$\pm$.28} & \textbf{.70$\pm$.03}  & $\uparrow$ .25 & $\downarrow$ .15\\
        \rowcolor{white!15} $\mathcal{V}_{16}$ & .22$\pm$.32 & .45$\pm$.24 & .32$\pm$.29 & .19$\pm$.27 & .36$\pm$.27 & .12$\pm$.20 & .11$\pm$.- - & \underline{.59$\pm$.23} & \textbf{.60$\pm$.27} & .53$\pm$.20  & $\uparrow$ .01 & -\\
        \hline
        \hline
        \end{tabular}
    }
\end{table*}

\vspace{0.5mm}
\noindent \emph{\textbf{Accuracy and Stability Analysis.}}
We give the performance comparison of training-based modes ($DRL_{re}$ and $DRL_{con}$) and testing-based modes ($DRL_{all}$ and $DRL_{one}$) of {\framework} with the baselines in Table \ref{tab:online_train} and Table \ref{tab:online_test}, respectively.
Due to the action space of {\framework} has the $stop$ action, it can automatically terminate the search.
To control the synchronization of baselines' experimental conditions, we use the average clustering rounds consumed when {\framework} is automatically terminated as the maximum round of baselines for the corresponding task ($30$ for Table \ref{tab:online_train} and $16$ for Table \ref{tab:online_test}).
The results show that the means of NMI scores of the training-based and testing-based search modes are improved by about $9\%$ and $9\%$ on average over multiple blocks, respectively, and the variances of performance are reduced by about $4\%$ and $2\%$, respectively.
Specifically, Table \ref{tab:online_train} firstly shows that similar to the offline tasks, $DRL_{re}$, which is based on re-training, still retains the significant advantage in online tasks.
Secondly, compared $DRL_{con}$ which is capable of continuous incremental learning with $DRL_{re}$, $DRL_{con}$ has a performance improvement of up to $14\%$ with a significant decrease in variance.
Third, from Table \ref{tab:online_test}, it can be found that in testing-based modes, $DRL_{all}$ and $DRL_{one}$  without labels (without reward function) can significantly exceed the baselines that require labels to establish the objective function.
Fourth, $DRL_{one}$ which is regularly maintained has higher performance and less variance than $DRL_{all}$.
These results demonstrate the capability of {\framework} to retain historical experience and the advantages of learnable DBSCAN parameter search for accuracy and stability.
In addition, although KDist can determine parameters without labels and iterations, its accuracy is relatively low.

\begin{figure}[t]
\centering
\includegraphics[width=8.5cm]{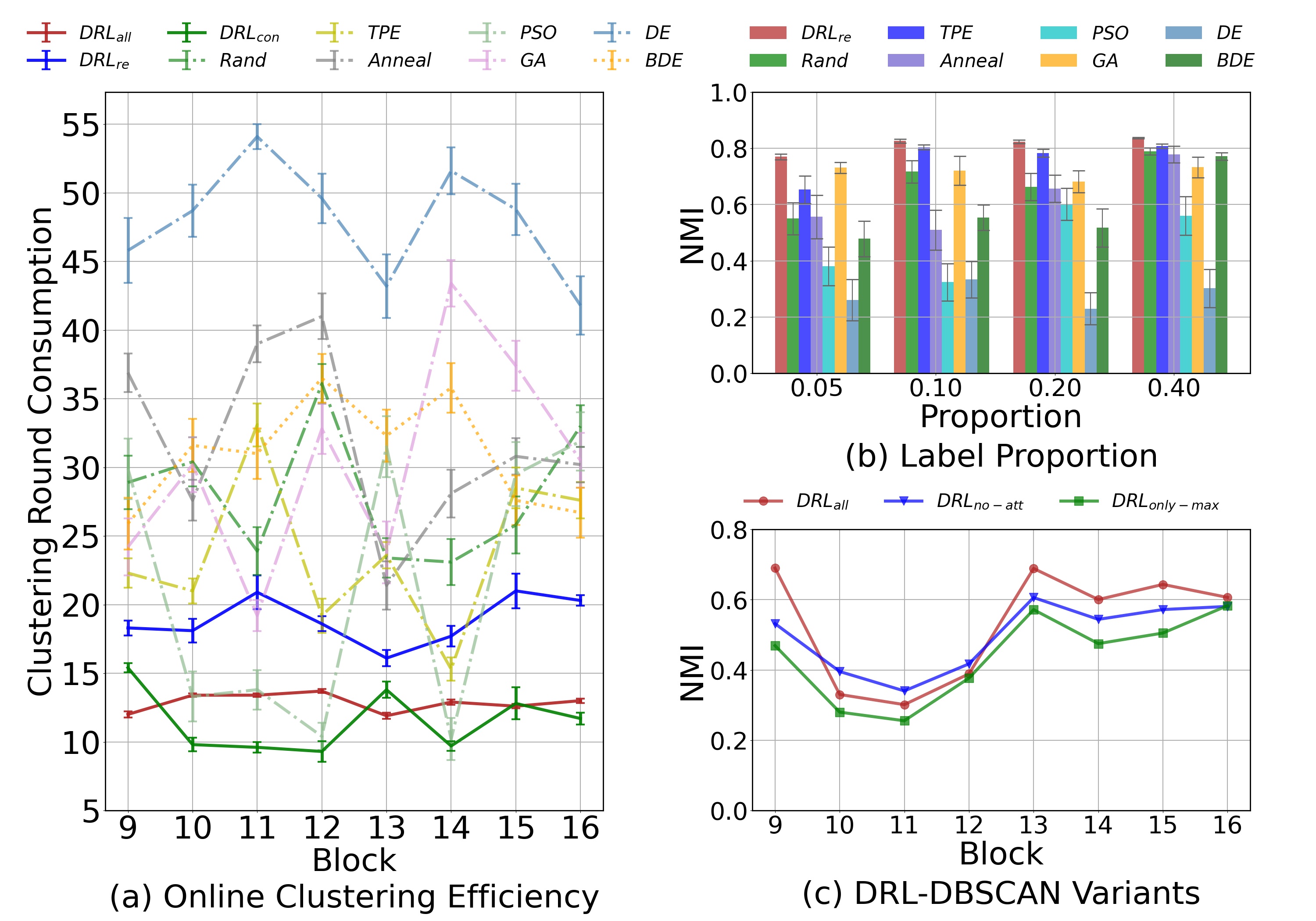}\vspace{-1em}
\centering
\caption{Comparison in online and offline tasks.}\label{fig:mix}
\vspace{-1.2mm}
\end{figure}

\begin{figure}
\centering
\includegraphics[width=8.5cm]{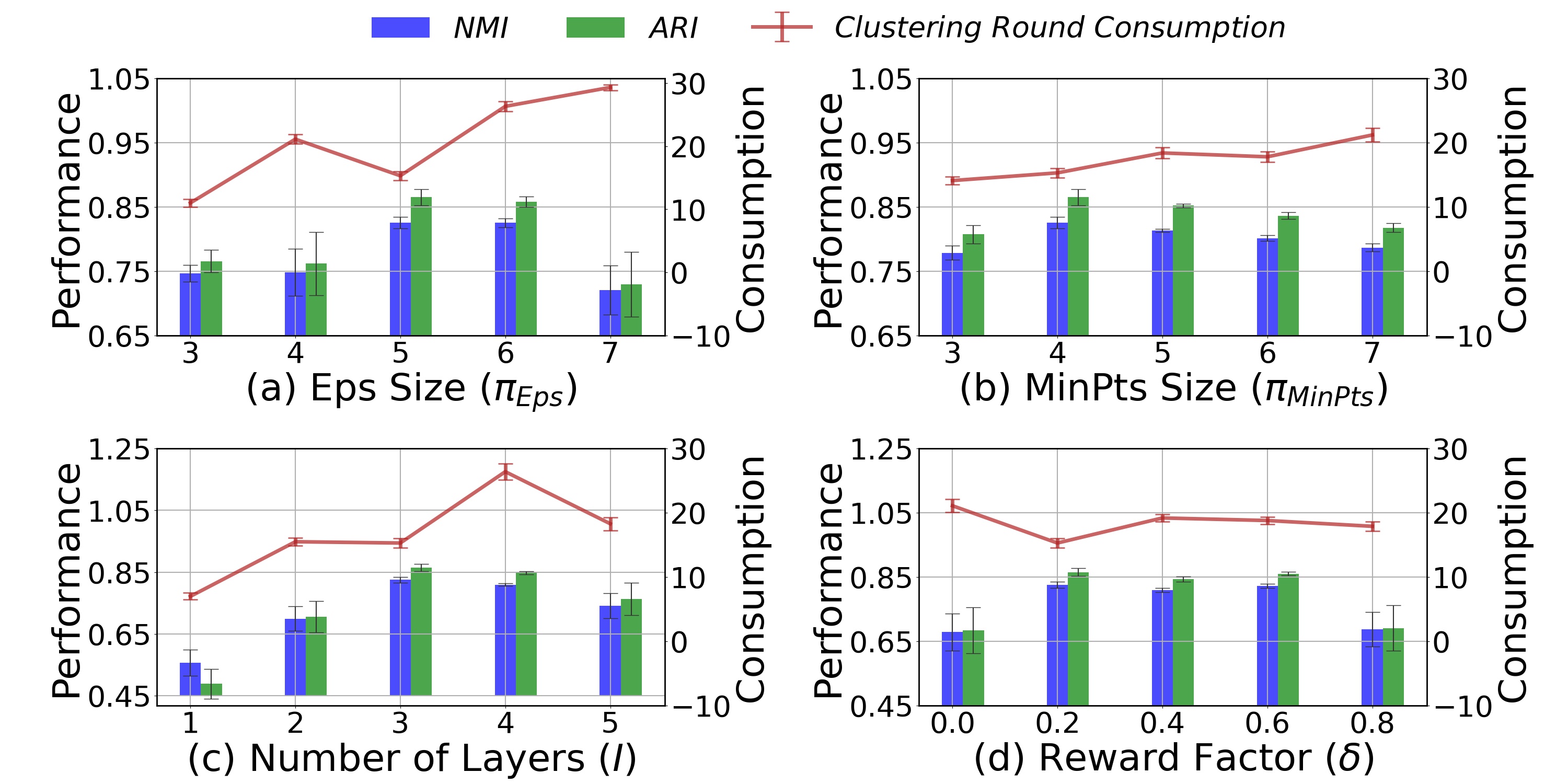}\vspace{-1em}
\centering
\caption{Parameter sensitivity.}\label{fig:hyperparameter}
\vspace{-3.2mm}
\end{figure}

\vspace{0.5mm}
\noindent \emph{\textbf{Efficiency Analysis.}}
$DRL_{all}$ and $DRL_{one}$ automatically search for the optimal DBSCAN parameters (end point parameters) without labels.
In order to better analyze these two testing-based parameter search modes, we compare the number of clustering rounds required of other methods to reach the NMI scores of the $DRL_{all}$ end-point parameters in the online tasks (Fig. \ref{fig:mix}(a)).
In the figure, the short vertical lines are the result variances.
We can see that $DRL_{all}$ reaches the optimal results within the consumption range of $11$-$14$ rounds, while other baselines require more rounds when reaching the corresponding NMI.
Moreover, many baselines' round consumption over different blocks fluctuates significantly, and the variance in the same block is also large.
The above observation suggests that the parameter search efficiency of {\framework}'s testing-based modes without labels exceeds that of the baselines which require labels.
Additionally, $DRL_{con}$ consumes fewer rounds than $DRL_{re}$ when reaching the same NMI, which also proves the advantage of {\framework}'s learning ability in terms of efficiency.

\vspace{0.5mm}
\noindent \emph{\textbf{{\framework} Variants.}}
To better evaluate the design of states and rewards in Sec. \ref{sec:parameter_search}, we compare two variants with $DRL_{all}$ in the online tasks, namely $DRL_{no-att}$ (state has no attention mechanism) and $DRL_{only-max}$ (reward only based on future maximum immediate reward).
The results in Fig. \ref{fig:mix}(c) show that the full structure of $DRL_{all}$ has better NMI scores than the variants, and brings the maximum performance increase of $0.16$, which represents the necessity of setting the local state and end point immediate reward.

\subsection{Hyperparameter Sensitivity}\label{sec:hyperparameter}
Fig. \ref{fig:hyperparameter} shows the results of the offline evaluation of $DRL_{re}$ on the Pathbased for four hyperparameters.
Fig. \ref{fig:hyperparameter}(a) and Fig. \ref{fig:hyperparameter}(b) compare a set of parameter space sizes of $Eps$ and $MinPts$ involved in the Eq. (\ref{eq:boundary}), respectively.
It can be found that the parameter space that is too large or too small for $Eps$ will cause performance loss and a decrease in search efficiency, while $MinPts$ is less sensitive to the parameter space's size change.
Fig. \ref{fig:hyperparameter}(c) analyzes the effect of different numbers of recursive layers on the search results. 
The results show that a suitable number of recurrent layers helps to obtain stable performance results.
It is worth noting that the number of layers does not require much tuning, as we use the early-stop mechanism described in Sec. \ref{sec:mode} to avoid overgrowing layers.
Fig. \ref{fig:hyperparameter}(d) compares the different influence weights of end point immediate reward and future maximum immediate reward on the final reward (Eq.\ref{eq:reward}).
The results show that equalizing the contribution of the two immediate rewards to the final reward can help improve the performance of the {\framework}.

%% file: 5-relatedwork.tex
\section{RELATED WORK}\label{sec:related_work}

\vspace{0.5mm}
\noindent \emph{\textbf{Automatic DBSCAN parameter determination.}}
\tobeupdated{DBSCAN is heavily dependent on two sensitive parameters ($Eps$ and $MinPts$) requiring prior knowledge for tuning.}
Numerous works propose different solutions for tuning the above.
OPTICS \cite{ankerst1999optics} is an extension of DBSCAN, which establishes cluster sorting based on reachability to obtain the $Eps$.
However, it needs to pre-determine the appropriate value of $MinPts$, and the acquisition of $Eps$ needs to interact with the user.
V-DBSCAN \cite{lu2007vdbscan} and KDDClus \cite{Mitra2011kddclus} plot the curve by using the sorted distance of any object to its $k$-th nearest object, and use the significant change on the curve as a series of candidate values for the $Eps$ parameter.
Similar methods include DSets-DBSCAN \cite{hou2016dsets}, Outlier \cite{akbari2016outlier} and RNN-DBSCAN \cite{bryant2017rnn_dbscan}, all of which require a fixed $MinPts$ value or a predetermined the number of nearest neighbors $k$, and the obtained candidate $Eps$ parameters may not be unique.
Beside the above work, there are some works \cite{darong2012grid, diao2018lpdbscsan} that consider combining DBSCAN with grid clustering to judge the density trend of raw samples according to the size and shape of each data region through pre-determined grid partition parameters.
Although these methods reduce the difficulty of parameter selection to a certain extent, they still require the user to decide at least one parameter heuristically, making them inflexible in changing data.

\vspace{0.5mm}
\noindent \emph{\textbf{Hyperparameter Optimization.}}
For the parameters of DBSCAN, another feasible parameter decision method is based on the Hyperparameter Optimization (HO) algorithm.
\tobeupdated{The classic HO methods are model-free methods, including grid search \cite{darong2012grid} that searches for all possible parameters, and random search \cite{bergstra2012rand} etc.}
\tobeupdated{Another approach is Bayesian optimization methods such as BO-TPE \cite{bergstra2011tpe}, SMAC \cite{hutter2011smac}, which optimize search efficiency using prior experience.}
In addition, meta-heuristic optimization methods, such as simulated annealing \cite{kirkpatrick1983anneal}, genetic \cite{lessmann2005ga}, particle swarm \cite{shi1998pso} and differential evolution \cite{qin2008de}, can solve non-convex, non-continuous and non-smooth optimization problems by simulating physical, biological and other processes to search \tobeupdated{\cite{yang2020hyperparameter}}.
Based on meta-heuristic optimization algorithms, some works propose HO methods for DBSCAN.
BDE-DBSCAN \cite{karami2014bdedbscan} targets an external purity index, selects $MinPts$ parameters based on a binary differential evolution algorithm, and selects $Eps$ parameters using a tournament selection algorithm.
MOGA-DBSCAN \cite{falahiazar2021moga_dbscan} proposes the outlier-index as a new internal index method for the objective function and selects parameters based on a multi-objective genetic algorithm.
Although HO methods avoid handcrafted heuristic decision parameters, they require an accurate objective function (clustering external/internal metrics) and cannot cope with the problem of unlabeled data and the error of internal metrics.
While {\framework} can not only perform DBSCAN clustering state-aware parametric search based on the objective function, but also retain the learned search experience and conduct searches without the objective function.

\vspace{0.5mm}
\noindent \emph{\textbf{Reinforcement Learning Clustering.}}
Recently, some works that intersect Reinforcement Learning (RL) and clustering algorithms have been proposed.
For example, MCTS Clustering \cite{brehmer2020rl_physics} in particle physics task builds high-quality hierarchical clusters through Monte Carlo tree search to reconstruct primitive elementary particles from observed final-state particles.
\cite{grua2018rl_health} which targets the health and medical domain leverages two clustering algorithms, and RL to cluster users who exhibit similar behaviors.
Both of these works are field-specific RL clustering methods. Compared with {\framework}, the Markov process they constructed is only applicable to fixed tasks, and is not a general clustering method.
Besides the above work, \cite{bagherjeiran2005rl_kmeans} proposes an improved K-Means clustering algorithm that selects the weights of distance metrics in different dimensions through RL.
Although this method effectively improves the performance of traditional K-Means, it needs to pre-determine the number of clusters $k$, which has limitations.

%% file: 6-conclusion.tex
\section{CONCLUSION}

In this paper, we propose an adaptive DBSCAN parameter search framework based on Deep Reinforcement Learning.
In the proposed {\framework} framework, the agents that modulate the parameter search direction by sensing the clustering environment are used to interact with the DBSCAN algorithm.
A recursive search mechanism is devised to avoid the search performance decline caused by a large parameter space.
The experimental results of the four working modes demonstrate that the proposed framework not only has high accuracy, stability and efficiency in searching parameters based on the objective function, but also maintains an effective performance when searching parameters without external incentives.

%% file: 0-main.bbl

\begin{thebibliography}{58}


\ifx \showCODEN    \undefined \def \showCODEN     #1{\unskip}     \fi
\ifx \showDOI      \undefined \def \showDOI       #1{#1}\fi
\ifx \showISBNx    \undefined \def \showISBNx     #1{\unskip}     \fi
\ifx \showISBNxiii \undefined \def \showISBNxiii  #1{\unskip}     \fi
\ifx \showISSN     \undefined \def \showISSN      #1{\unskip}     \fi
\ifx \showLCCN     \undefined \def \showLCCN      #1{\unskip}     \fi
\ifx \shownote     \undefined \def \shownote      #1{#1}          \fi
\ifx \showarticletitle \undefined \def \showarticletitle #1{#1}   \fi
\ifx \showURL      \undefined \def \showURL       {\relax}        \fi
\providecommand\bibfield[2]{#2}
\providecommand\bibinfo[2]{#2}
\providecommand\natexlab[1]{#1}
\providecommand\showeprint[2][]{arXiv:#2}

\bibitem[\protect\citeauthoryear{??}{_20}{2022}]%
        {_2022scikitopt}
 \bibinfo{year}{2022}\natexlab{}.
\newblock \bibinfo{title}{scikit-opt}.
\newblock
\newblock
\urldef\tempurl%
\url{https://github.com/guofei9987/scikit-opt}
\showURL{%
\tempurl}


\bibitem[\protect\citeauthoryear{Akbari and Unland}{Akbari and Unland}{2016}]%
        {akbari2016outlier}
\bibfield{author}{\bibinfo{person}{Zohreh Akbari} {and} \bibinfo{person}{Rainer
  Unland}.} \bibinfo{year}{2016}\natexlab{}.
\newblock \showarticletitle{Automated determination of the input parameter of
  DBSCAN based on outlier detection}. In \bibinfo{booktitle}{\emph{Ifip
  international conference on artificial intelligence applications and
  innovations}}. Springer, \bibinfo{pages}{280--291}.
\newblock


\bibitem[\protect\citeauthoryear{Ankerst, Breunig, Kriegel, and Sander}{Ankerst
  et~al\mbox{.}}{1999}]%
        {ankerst1999optics}
\bibfield{author}{\bibinfo{person}{Mihael Ankerst}, \bibinfo{person}{Markus~M
  Breunig}, \bibinfo{person}{Hans-Peter Kriegel}, {and}
  \bibinfo{person}{J{\"o}rg Sander}.} \bibinfo{year}{1999}\natexlab{}.
\newblock \showarticletitle{OPTICS: Ordering points to identify the clustering
  structure}.
\newblock \bibinfo{journal}{\emph{ACM Sigmod record}} \bibinfo{volume}{28},
  \bibinfo{number}{2} (\bibinfo{year}{1999}), \bibinfo{pages}{49--60}.
\newblock


\bibitem[\protect\citeauthoryear{Bagherjeiran, Eick, and Vilalta}{Bagherjeiran
  et~al\mbox{.}}{2005}]%
        {bagherjeiran2005rl_kmeans}
\bibfield{author}{\bibinfo{person}{Abraham Bagherjeiran},
  \bibinfo{person}{Christoph~F Eick}, {and} \bibinfo{person}{Ricardo Vilalta}.}
  \bibinfo{year}{2005}\natexlab{}.
\newblock \showarticletitle{Adaptive clustering: Better representatives with
  reinforcement learning}.
\newblock \bibinfo{journal}{\emph{Department of Computer Science, University of
  Houston, Houston}} (\bibinfo{year}{2005}).
\newblock


\bibitem[\protect\citeauthoryear{Bergstra, Bardenet, Bengio, and
  K{\'e}gl}{Bergstra et~al\mbox{.}}{2011}]%
        {bergstra2011tpe}
\bibfield{author}{\bibinfo{person}{James Bergstra}, \bibinfo{person}{R{\'e}mi
  Bardenet}, \bibinfo{person}{Yoshua Bengio}, {and} \bibinfo{person}{Bal{\'a}zs
  K{\'e}gl}.} \bibinfo{year}{2011}\natexlab{}.
\newblock \showarticletitle{Algorithms for hyper-parameter optimization}.
\newblock \bibinfo{journal}{\emph{Advances in neural information processing
  systems}}  \bibinfo{volume}{24} (\bibinfo{year}{2011}).
\newblock


\bibitem[\protect\citeauthoryear{Bergstra and Bengio}{Bergstra and
  Bengio}{2012}]%
        {bergstra2012rand}
\bibfield{author}{\bibinfo{person}{James Bergstra} {and}
  \bibinfo{person}{Yoshua Bengio}.} \bibinfo{year}{2012}\natexlab{}.
\newblock \showarticletitle{Random search for hyper-parameter optimization.}
\newblock \bibinfo{journal}{\emph{Journal of machine learning research}}
  \bibinfo{volume}{13}, \bibinfo{number}{2} (\bibinfo{year}{2012}).
\newblock


\bibitem[\protect\citeauthoryear{Bergstra, Yamins, and Cox}{Bergstra
  et~al\mbox{.}}{2013}]%
        {bergstra2013hyperopt}
\bibfield{author}{\bibinfo{person}{James Bergstra}, \bibinfo{person}{Daniel
  Yamins}, {and} \bibinfo{person}{David Cox}.} \bibinfo{year}{2013}\natexlab{}.
\newblock \showarticletitle{Making a science of model search: Hyperparameter
  optimization in hundreds of dimensions for vision architectures}. In
  \bibinfo{booktitle}{\emph{International conference on machine learning}}.
  PMLR, \bibinfo{pages}{115--123}.
\newblock


\bibitem[\protect\citeauthoryear{Bom, Henken, and Wiering}{Bom
  et~al\mbox{.}}{2013}]%
        {bom2013pac_man}
\bibfield{author}{\bibinfo{person}{Luuk Bom}, \bibinfo{person}{Ruud Henken},
  {and} \bibinfo{person}{Marco Wiering}.} \bibinfo{year}{2013}\natexlab{}.
\newblock \showarticletitle{Reinforcement learning to train Ms. Pac-Man using
  higher-order action-relative inputs}. In \bibinfo{booktitle}{\emph{2013 IEEE
  Symposium on Adaptive Dynamic Programming and Reinforcement Learning
  (ADPRL)}}. IEEE, \bibinfo{pages}{156--163}.
\newblock


\bibitem[\protect\citeauthoryear{Brehmer, Macaluso, Pappadopulo, and
  Cranmer}{Brehmer et~al\mbox{.}}{2020}]%
        {brehmer2020rl_physics}
\bibfield{author}{\bibinfo{person}{Johann Brehmer}, \bibinfo{person}{Sebastian
  Macaluso}, \bibinfo{person}{Duccio Pappadopulo}, {and} \bibinfo{person}{Kyle
  Cranmer}.} \bibinfo{year}{2020}\natexlab{}.
\newblock \showarticletitle{Hierarchical clustering in particle physics through
  reinforcement learning}.
\newblock \bibinfo{journal}{\emph{arXiv preprint arXiv:2011.08191}}
  (\bibinfo{year}{2020}).
\newblock


\bibitem[\protect\citeauthoryear{Bryant and Cios}{Bryant and Cios}{2017}]%
        {bryant2017rnn_dbscan}
\bibfield{author}{\bibinfo{person}{Avory Bryant} {and}
  \bibinfo{person}{Krzysztof Cios}.} \bibinfo{year}{2017}\natexlab{}.
\newblock \showarticletitle{RNN-DBSCAN: A density-based clustering algorithm
  using reverse nearest neighbor density estimates}.
\newblock \bibinfo{journal}{\emph{IEEE Transactions on Knowledge and Data
  Engineering}} \bibinfo{volume}{30}, \bibinfo{number}{6}
  (\bibinfo{year}{2017}), \bibinfo{pages}{1109--1121}.
\newblock


\bibitem[\protect\citeauthoryear{Chang and Yeung}{Chang and Yeung}{2008}]%
        {chang2008pathbased}
\bibfield{author}{\bibinfo{person}{Hong Chang} {and} \bibinfo{person}{Dit-Yan
  Yeung}.} \bibinfo{year}{2008}\natexlab{}.
\newblock \showarticletitle{Robust path-based spectral clustering}.
\newblock \bibinfo{journal}{\emph{Pattern Recognition}} \bibinfo{volume}{41},
  \bibinfo{number}{1} (\bibinfo{year}{2008}), \bibinfo{pages}{191--203}.
\newblock


\bibitem[\protect\citeauthoryear{Darong and Peng}{Darong and Peng}{2012}]%
        {darong2012grid}
\bibfield{author}{\bibinfo{person}{Huang Darong} {and} \bibinfo{person}{Wang
  Peng}.} \bibinfo{year}{2012}\natexlab{}.
\newblock \showarticletitle{Grid-based DBSCAN algorithm with referential
  parameters}.
\newblock \bibinfo{journal}{\emph{Physics Procedia}}  \bibinfo{volume}{24}
  (\bibinfo{year}{2012}), \bibinfo{pages}{1166--1170}.
\newblock


\bibitem[\protect\citeauthoryear{Diao, Liang, and Fan}{Diao
  et~al\mbox{.}}{2018}]%
        {diao2018lpdbscsan}
\bibfield{author}{\bibinfo{person}{Kejing Diao}, \bibinfo{person}{Yongquan
  Liang}, {and} \bibinfo{person}{Jiancong Fan}.}
  \bibinfo{year}{2018}\natexlab{}.
\newblock \showarticletitle{An improved DBSCAN algorithm using local
  parameters}. In \bibinfo{booktitle}{\emph{International CCF Conference on
  Artificial Intelligence}}. Springer, \bibinfo{pages}{3--12}.
\newblock


\bibitem[\protect\citeauthoryear{Dulac-Arnold, Evans, van Hasselt, Sunehag,
  Lillicrap, Hunt, Mann, Weber, Degris, and Coppin}{Dulac-Arnold
  et~al\mbox{.}}{2015}]%
        {dulacarnold2016large_discrete}
\bibfield{author}{\bibinfo{person}{Gabriel Dulac-Arnold},
  \bibinfo{person}{Richard Evans}, \bibinfo{person}{Hado van Hasselt},
  \bibinfo{person}{Peter Sunehag}, \bibinfo{person}{Timothy Lillicrap},
  \bibinfo{person}{Jonathan Hunt}, \bibinfo{person}{Timothy Mann},
  \bibinfo{person}{Theophane Weber}, \bibinfo{person}{Thomas Degris}, {and}
  \bibinfo{person}{Ben Coppin}.} \bibinfo{year}{2015}\natexlab{}.
\newblock \showarticletitle{Deep reinforcement learning in large discrete
  action spaces}.
\newblock \bibinfo{journal}{\emph{arXiv preprint arXiv:1512.07679}}
  (\bibinfo{year}{2015}).
\newblock


\bibitem[\protect\citeauthoryear{Ester, Kriegel, Sander, Xu,
  et~al\mbox{.}}{Ester et~al\mbox{.}}{1996}]%
        {ester1996dbscan}
\bibfield{author}{\bibinfo{person}{Martin Ester}, \bibinfo{person}{Hans-Peter
  Kriegel}, \bibinfo{person}{J{\"o}rg Sander}, \bibinfo{person}{Xiaowei Xu},
  {et~al\mbox{.}}} \bibinfo{year}{1996}\natexlab{}.
\newblock \showarticletitle{A density-based algorithm for discovering clusters
  in large spatial databases with noise.}. In \bibinfo{booktitle}{\emph{kdd}},
  Vol.~\bibinfo{volume}{96}. \bibinfo{pages}{226--231}.
\newblock


\bibitem[\protect\citeauthoryear{Est{\'e}vez, Tesmer, Perez, and
  Zurada}{Est{\'e}vez et~al\mbox{.}}{2009}]%
        {estevez2009nmi}
\bibfield{author}{\bibinfo{person}{Pablo~A Est{\'e}vez},
  \bibinfo{person}{Michel Tesmer}, \bibinfo{person}{Claudio~A Perez}, {and}
  \bibinfo{person}{Jacek~M Zurada}.} \bibinfo{year}{2009}\natexlab{}.
\newblock \showarticletitle{Normalized mutual information feature selection}.
\newblock \bibinfo{journal}{\emph{IEEE Transactions on neural networks}}
  \bibinfo{volume}{20}, \bibinfo{number}{2} (\bibinfo{year}{2009}),
  \bibinfo{pages}{189--201}.
\newblock


\bibitem[\protect\citeauthoryear{Falahiazar, Bagheri, and Reshadi}{Falahiazar
  et~al\mbox{.}}{2021}]%
        {falahiazar2021moga_dbscan}
\bibfield{author}{\bibinfo{person}{Zeinab Falahiazar}, \bibinfo{person}{Alireza
  Bagheri}, {and} \bibinfo{person}{Midia Reshadi}.}
  \bibinfo{year}{2021}\natexlab{}.
\newblock \showarticletitle{Determining the Parameters of DBSCAN Automatically
  Using the Multi-Objective Genetic Algorithm.}
\newblock \bibinfo{journal}{\emph{J. Inf. Sci. Eng.}} \bibinfo{volume}{37},
  \bibinfo{number}{1} (\bibinfo{year}{2021}), \bibinfo{pages}{157--183}.
\newblock


\bibitem[\protect\citeauthoryear{Fan, Guo, and Ren}{Fan et~al\mbox{.}}{2021}]%
        {fan2021consumer}
\bibfield{author}{\bibinfo{person}{Tianhui Fan}, \bibinfo{person}{Naijing Guo},
  {and} \bibinfo{person}{Yujie Ren}.} \bibinfo{year}{2021}\natexlab{}.
\newblock \showarticletitle{Consumer clusters detection with geo-tagged social
  network data using DBSCAN algorithm: a case study of the Pearl River Delta in
  China}.
\newblock \bibinfo{journal}{\emph{GeoJournal}} \bibinfo{volume}{86},
  \bibinfo{number}{1} (\bibinfo{year}{2021}), \bibinfo{pages}{317--337}.
\newblock


\bibitem[\protect\citeauthoryear{Fan and Xu}{Fan and Xu}{2019}]%
        {fan2019seismic_data}
\bibfield{author}{\bibinfo{person}{Z Fan} {and} \bibinfo{person}{Xiaolong Xu}.}
  \bibinfo{year}{2019}\natexlab{}.
\newblock \showarticletitle{Application and visualization of typical clustering
  algorithms in seismic data analysis}.
\newblock \bibinfo{journal}{\emph{Procedia Computer Science}}
  \bibinfo{volume}{151} (\bibinfo{year}{2019}), \bibinfo{pages}{171--178}.
\newblock


\bibitem[\protect\citeauthoryear{Francis, Villagrasa, and Clairand}{Francis
  et~al\mbox{.}}{2011}]%
        {francis2011dna_damage}
\bibfield{author}{\bibinfo{person}{Ziad Francis}, \bibinfo{person}{Carmen
  Villagrasa}, {and} \bibinfo{person}{Isabelle Clairand}.}
  \bibinfo{year}{2011}\natexlab{}.
\newblock \showarticletitle{Simulation of DNA damage clustering after proton
  irradiation using an adapted DBSCAN algorithm}.
\newblock \bibinfo{journal}{\emph{Computer methods and programs in
  biomedicine}} \bibinfo{volume}{101}, \bibinfo{number}{3}
  (\bibinfo{year}{2011}), \bibinfo{pages}{265--270}.
\newblock


\bibitem[\protect\citeauthoryear{Fr{\"a}nti and Sieranoja}{Fr{\"a}nti and
  Sieranoja}{2018}]%
        {Pasi2018benchmark}
\bibfield{author}{\bibinfo{person}{Pasi Fr{\"a}nti} {and} \bibinfo{person}{Sami
  Sieranoja}.} \bibinfo{year}{2018}\natexlab{}.
\newblock \showarticletitle{K-means properties on six clustering benchmark
  datasets}.
\newblock \bibinfo{journal}{\emph{Applied intelligence}} \bibinfo{volume}{48},
  \bibinfo{number}{12} (\bibinfo{year}{2018}), \bibinfo{pages}{4743--4759}.
\newblock


\bibitem[\protect\citeauthoryear{Fujimoto, Hoof, and Meger}{Fujimoto
  et~al\mbox{.}}{2018}]%
        {scott2018td3}
\bibfield{author}{\bibinfo{person}{Scott Fujimoto}, \bibinfo{person}{Herke
  Hoof}, {and} \bibinfo{person}{David Meger}.} \bibinfo{year}{2018}\natexlab{}.
\newblock \showarticletitle{Addressing function approximation error in
  actor-critic methods}. In \bibinfo{booktitle}{\emph{International conference
  on machine learning}}. PMLR, \bibinfo{pages}{1587--1596}.
\newblock


\bibitem[\protect\citeauthoryear{Gionis, Mannila, and Tsaparas}{Gionis
  et~al\mbox{.}}{2007}]%
        {gionis2007aggregation}
\bibfield{author}{\bibinfo{person}{Aristides Gionis}, \bibinfo{person}{Heikki
  Mannila}, {and} \bibinfo{person}{Panayiotis Tsaparas}.}
  \bibinfo{year}{2007}\natexlab{}.
\newblock \showarticletitle{Clustering aggregation}.
\newblock \bibinfo{journal}{\emph{Acm transactions on knowledge discovery from
  data (tkdd)}} \bibinfo{volume}{1}, \bibinfo{number}{1}
  (\bibinfo{year}{2007}), \bibinfo{pages}{4--es}.
\newblock


\bibitem[\protect\citeauthoryear{Grua and Hoogendoorn}{Grua and
  Hoogendoorn}{2018}]%
        {grua2018rl_health}
\bibfield{author}{\bibinfo{person}{Eoin~Martino Grua} {and}
  \bibinfo{person}{Mark Hoogendoorn}.} \bibinfo{year}{2018}\natexlab{}.
\newblock \showarticletitle{Exploring clustering techniques for effective
  reinforcement learning based personalization for health and wellbeing}. In
  \bibinfo{booktitle}{\emph{2018 IEEE Symposium Series on Computational
  Intelligence (SSCI)}}. IEEE, \bibinfo{pages}{813--820}.
\newblock


\bibitem[\protect\citeauthoryear{Guan, Fung, and Yue}{Guan
  et~al\mbox{.}}{2018}]%
        {guan2018social_tagging}
\bibfield{author}{\bibinfo{person}{Chun Guan}, \bibinfo{person}{Yuen Kevin~Kam
  Fung}, {and} \bibinfo{person}{Yong Yue}.} \bibinfo{year}{2018}\natexlab{}.
\newblock \showarticletitle{Towards a Personalized Item Recommendation Approach
  in Social Tagging Systems Using Intuitionistic Fuzzy DBSCAN}. In
  \bibinfo{booktitle}{\emph{2018 10th International Conference on Intelligent
  Human-Machine Systems and Cybernetics (IHMSC)}}, Vol.~\bibinfo{volume}{1}.
  IEEE, \bibinfo{pages}{361--364}.
\newblock


\bibitem[\protect\citeauthoryear{Hou, Gao, and Li}{Hou et~al\mbox{.}}{2016}]%
        {hou2016dsets}
\bibfield{author}{\bibinfo{person}{Jian Hou}, \bibinfo{person}{Huijun Gao},
  {and} \bibinfo{person}{Xuelong Li}.} \bibinfo{year}{2016}\natexlab{}.
\newblock \showarticletitle{DSets-DBSCAN: A parameter-free clustering
  algorithm}.
\newblock \bibinfo{journal}{\emph{IEEE Transactions on Image Processing}}
  \bibinfo{volume}{25}, \bibinfo{number}{7} (\bibinfo{year}{2016}),
  \bibinfo{pages}{3182--3193}.
\newblock


\bibitem[\protect\citeauthoryear{Huang, Bao, Zhang, and Feng}{Huang
  et~al\mbox{.}}{2019}]%
        {huang2019time_series}
\bibfield{author}{\bibinfo{person}{Mengxing Huang}, \bibinfo{person}{Qili Bao},
  \bibinfo{person}{Yu Zhang}, {and} \bibinfo{person}{Wenlong Feng}.}
  \bibinfo{year}{2019}\natexlab{}.
\newblock \showarticletitle{A hybrid algorithm for forecasting financial time
  series data based on DBSCAN and SVR}.
\newblock \bibinfo{journal}{\emph{Information}} \bibinfo{volume}{10},
  \bibinfo{number}{3} (\bibinfo{year}{2019}), \bibinfo{pages}{103}.
\newblock


\bibitem[\protect\citeauthoryear{Hutter, Hoos, and Leyton-Brown}{Hutter
  et~al\mbox{.}}{2011}]%
        {hutter2011smac}
\bibfield{author}{\bibinfo{person}{Frank Hutter}, \bibinfo{person}{Holger~H
  Hoos}, {and} \bibinfo{person}{Kevin Leyton-Brown}.}
  \bibinfo{year}{2011}\natexlab{}.
\newblock \showarticletitle{Sequential model-based optimization for general
  algorithm configuration}. In \bibinfo{booktitle}{\emph{International
  conference on learning and intelligent optimization}}. Springer,
  \bibinfo{pages}{507--523}.
\newblock


\bibitem[\protect\citeauthoryear{Kanervisto, Scheller, and
  Hautam{\"a}ki}{Kanervisto et~al\mbox{.}}{2020}]%
        {Anssi2020large_continuous}
\bibfield{author}{\bibinfo{person}{Anssi Kanervisto},
  \bibinfo{person}{Christian Scheller}, {and} \bibinfo{person}{Ville
  Hautam{\"a}ki}.} \bibinfo{year}{2020}\natexlab{}.
\newblock \showarticletitle{Action space shaping in deep reinforcement
  learning}. In \bibinfo{booktitle}{\emph{2020 IEEE Conference on Games
  (CoG)}}. IEEE, \bibinfo{pages}{479--486}.
\newblock


\bibitem[\protect\citeauthoryear{Karami and Johansson}{Karami and
  Johansson}{2014}]%
        {karami2014bdedbscan}
\bibfield{author}{\bibinfo{person}{Amin Karami} {and} \bibinfo{person}{Ronnie
  Johansson}.} \bibinfo{year}{2014}\natexlab{}.
\newblock \showarticletitle{Choosing DBSCAN parameters automatically using
  differential evolution}.
\newblock \bibinfo{journal}{\emph{International Journal of Computer
  Applications}} \bibinfo{volume}{91}, \bibinfo{number}{7}
  (\bibinfo{year}{2014}), \bibinfo{pages}{1--11}.
\newblock


\bibitem[\protect\citeauthoryear{Kazemi-Beydokhti, Ali~Abbaspour, and
  Mojarab}{Kazemi-Beydokhti et~al\mbox{.}}{2017}]%
        {kazemi2017iran}
\bibfield{author}{\bibinfo{person}{Mohammad Kazemi-Beydokhti},
  \bibinfo{person}{Rahim Ali~Abbaspour}, {and} \bibinfo{person}{Masoud
  Mojarab}.} \bibinfo{year}{2017}\natexlab{}.
\newblock \showarticletitle{Spatio-Temporal Modeling of Seismic Provinces of
  Iran Using DBSCAN Algorithm}.
\newblock \bibinfo{journal}{\emph{Pure \& Applied Geophysics}}
  \bibinfo{volume}{174}, \bibinfo{number}{5} (\bibinfo{year}{2017}).
\newblock


\bibitem[\protect\citeauthoryear{Kirkpatrick, Gelatt~Jr, and
  Vecchi}{Kirkpatrick et~al\mbox{.}}{1983}]%
        {kirkpatrick1983anneal}
\bibfield{author}{\bibinfo{person}{Scott Kirkpatrick},
  \bibinfo{person}{C~Daniel Gelatt~Jr}, {and} \bibinfo{person}{Mario~P
  Vecchi}.} \bibinfo{year}{1983}\natexlab{}.
\newblock \showarticletitle{Optimization by simulated annealing}.
\newblock \bibinfo{journal}{\emph{science}} \bibinfo{volume}{220},
  \bibinfo{number}{4598} (\bibinfo{year}{1983}), \bibinfo{pages}{671--680}.
\newblock


\bibitem[\protect\citeauthoryear{Konda and Tsitsiklis}{Konda and
  Tsitsiklis}{2000}]%
        {konda2000actor_critic}
\bibfield{author}{\bibinfo{person}{Vijay~R Konda} {and} \bibinfo{person}{John~N
  Tsitsiklis}.} \bibinfo{year}{2000}\natexlab{}.
\newblock \showarticletitle{Actor-critic algorithms}. In
  \bibinfo{booktitle}{\emph{Proceedings of the NIPS}}.
  \bibinfo{pages}{1008--1014}.
\newblock


\bibitem[\protect\citeauthoryear{Ku{\.z}elewska and Wichowski}{Ku{\.z}elewska
  and Wichowski}{2015}]%
        {kuzelewska2015music}
\bibfield{author}{\bibinfo{person}{Urszula Ku{\.z}elewska} {and}
  \bibinfo{person}{Krzysztof Wichowski}.} \bibinfo{year}{2015}\natexlab{}.
\newblock \showarticletitle{A modified clustering algorithm DBSCAN used in a
  collaborative filtering recommender system for music recommendation}. In
  \bibinfo{booktitle}{\emph{International conference on dependability and
  complex systems}}. Springer, \bibinfo{pages}{245--254}.
\newblock


\bibitem[\protect\citeauthoryear{Lai, Zhou, Hu, Bian, and Song}{Lai
  et~al\mbox{.}}{2019}]%
        {lai2019mvodbscan}
\bibfield{author}{\bibinfo{person}{Wenhao Lai}, \bibinfo{person}{Mengran Zhou},
  \bibinfo{person}{Feng Hu}, \bibinfo{person}{Kai Bian}, {and}
  \bibinfo{person}{Qi Song}.} \bibinfo{year}{2019}\natexlab{}.
\newblock \showarticletitle{A new DBSCAN parameters determination method based
  on improved MVO}.
\newblock \bibinfo{journal}{\emph{Ieee Access}}  \bibinfo{volume}{7}
  (\bibinfo{year}{2019}), \bibinfo{pages}{104085--104095}.
\newblock


\bibitem[\protect\citeauthoryear{Lessmann, Stahlbock, and Crone}{Lessmann
  et~al\mbox{.}}{2005}]%
        {lessmann2005ga}
\bibfield{author}{\bibinfo{person}{Stefan Lessmann}, \bibinfo{person}{Robert
  Stahlbock}, {and} \bibinfo{person}{Sven~F Crone}.}
  \bibinfo{year}{2005}\natexlab{}.
\newblock \showarticletitle{Optimizing hyperparameters of support vector
  machines by genetic algorithms.}. In \bibinfo{booktitle}{\emph{IC-AI}}.
  \bibinfo{pages}{74--82}.
\newblock


\bibitem[\protect\citeauthoryear{Li and Li}{Li and Li}{2007}]%
        {li2007public_facility}
\bibfield{author}{\bibinfo{person}{Xinyan Li} {and} \bibinfo{person}{Deren
  Li}.} \bibinfo{year}{2007}\natexlab{}.
\newblock \showarticletitle{Discovery of rules in urban public facility
  distribution based on DBSCAN clustering algorithm}. In
  \bibinfo{booktitle}{\emph{MIPPR 2007: Remote Sensing and GIS Data Processing
  and Applications; and Innovative Multispectral Technology and Applications}},
  Vol.~\bibinfo{volume}{6790}. International Society for Optics and Photonics,
  \bibinfo{pages}{67902E}.
\newblock


\bibitem[\protect\citeauthoryear{Lillicrap, Hunt, Pritzel, Heess, Erez, Tassa,
  Silver, and Wierstra}{Lillicrap et~al\mbox{.}}{2015}]%
        {lillicrap2015ddpg}
\bibfield{author}{\bibinfo{person}{Timothy~P Lillicrap},
  \bibinfo{person}{Jonathan~J Hunt}, \bibinfo{person}{Alexander Pritzel},
  \bibinfo{person}{Nicolas Heess}, \bibinfo{person}{Tom Erez},
  \bibinfo{person}{Yuval Tassa}, \bibinfo{person}{David Silver}, {and}
  \bibinfo{person}{Daan Wierstra}.} \bibinfo{year}{2015}\natexlab{}.
\newblock \showarticletitle{Continuous control with deep reinforcement
  learning}.
\newblock \bibinfo{journal}{\emph{arXiv preprint arXiv:1509.02971}}
  (\bibinfo{year}{2015}).
\newblock


\bibitem[\protect\citeauthoryear{Liu, Zhou, and Wu}{Liu et~al\mbox{.}}{2007}]%
        {lu2007vdbscan}
\bibfield{author}{\bibinfo{person}{Peng Liu}, \bibinfo{person}{Dong Zhou},
  {and} \bibinfo{person}{Naijun Wu}.} \bibinfo{year}{2007}\natexlab{}.
\newblock \showarticletitle{VDBSCAN: varied density based spatial clustering of
  applications with noise}. In \bibinfo{booktitle}{\emph{2007 International
  conference on service systems and service management}}. IEEE,
  \bibinfo{pages}{1--4}.
\newblock


\bibitem[\protect\citeauthoryear{Mitra and Nandy}{Mitra and Nandy}{2011}]%
        {Mitra2011kddclus}
\bibfield{author}{\bibinfo{person}{S. Mitra} {and} \bibinfo{person}{Jay
  Nandy}.} \bibinfo{year}{2011}\natexlab{}.
\newblock \showarticletitle{KDDClus : A Simple Method for Multi-Density
  Clustering}. In \bibinfo{booktitle}{\emph{Proceedings of International
  Workshop on Soft Computing Applications and Knowledge Discovery (SCAKD)}}.
  \bibinfo{pages}{72--76}.
\newblock


\bibitem[\protect\citeauthoryear{Mohammed, Cawthorne, and Abdulazeez}{Mohammed
  et~al\mbox{.}}{2018}]%
        {mohammed2018genes_patterns}
\bibfield{author}{\bibinfo{person}{Nwayyin~Najat Mohammed},
  \bibinfo{person}{Micheal Cawthorne}, {and} \bibinfo{person}{Adnan~Mohsin
  Abdulazeez}.} \bibinfo{year}{2018}\natexlab{}.
\newblock \showarticletitle{Detection of Genes Patterns with an Enhanced
  Partitioning-Based DBSCAN Algorithm}.
\newblock \bibinfo{journal}{\emph{Journal of information and communication
  engineering}} \bibinfo{volume}{4}, \bibinfo{number}{1}
  (\bibinfo{year}{2018}), \bibinfo{pages}{188--195}.
\newblock


\bibitem[\protect\citeauthoryear{Mundhenk, Goldsmith, Lusena, and
  Allender}{Mundhenk et~al\mbox{.}}{2000}]%
        {mundhenk2000mdp}
\bibfield{author}{\bibinfo{person}{Martin Mundhenk}, \bibinfo{person}{Judy
  Goldsmith}, \bibinfo{person}{Christopher Lusena}, {and} \bibinfo{person}{Eric
  Allender}.} \bibinfo{year}{2000}\natexlab{}.
\newblock \showarticletitle{Complexity of finite-horizon Markov decision
  process problems}.
\newblock \bibinfo{journal}{\emph{Journal of the ACM (JACM)}}
  \bibinfo{volume}{47}, \bibinfo{number}{4} (\bibinfo{year}{2000}),
  \bibinfo{pages}{681--720}.
\newblock


\bibitem[\protect\citeauthoryear{Pavlis, Dolega, and Singleton}{Pavlis
  et~al\mbox{.}}{2018}]%
        {pavlis2018retail_center}
\bibfield{author}{\bibinfo{person}{Michalis Pavlis}, \bibinfo{person}{Les
  Dolega}, {and} \bibinfo{person}{Alex Singleton}.}
  \bibinfo{year}{2018}\natexlab{}.
\newblock \showarticletitle{A modified DBSCAN clustering method to estimate
  retail center extent}.
\newblock \bibinfo{journal}{\emph{Geographical Analysis}} \bibinfo{volume}{50},
  \bibinfo{number}{2} (\bibinfo{year}{2018}), \bibinfo{pages}{141--161}.
\newblock


\bibitem[\protect\citeauthoryear{Peng, Zhang, Li, Cao, Pan, and Yu}{Peng
  et~al\mbox{.}}{2022}]%
        {peng2022reinforced}
\bibfield{author}{\bibinfo{person}{Hao Peng}, \bibinfo{person}{Ruitong Zhang},
  \bibinfo{person}{Shaoning Li}, \bibinfo{person}{Yuwei Cao},
  \bibinfo{person}{Shirui Pan}, {and} \bibinfo{person}{Philip Yu}.}
  \bibinfo{year}{2022}\natexlab{}.
\newblock \showarticletitle{Reinforced, incremental and cross-lingual event
  detection from social messages}.
\newblock \bibinfo{journal}{\emph{IEEE Transactions on Pattern Analysis and
  Machine Intelligence}} (\bibinfo{year}{2022}).
\newblock


\bibitem[\protect\citeauthoryear{Qin, Huang, and Suganthan}{Qin
  et~al\mbox{.}}{2008}]%
        {qin2008de}
\bibfield{author}{\bibinfo{person}{A~Kai Qin}, \bibinfo{person}{Vicky~Ling
  Huang}, {and} \bibinfo{person}{Ponnuthurai~N Suganthan}.}
  \bibinfo{year}{2008}\natexlab{}.
\newblock \showarticletitle{Differential evolution algorithm with strategy
  adaptation for global numerical optimization}.
\newblock \bibinfo{journal}{\emph{IEEE transactions on Evolutionary
  Computation}} \bibinfo{volume}{13}, \bibinfo{number}{2}
  (\bibinfo{year}{2008}), \bibinfo{pages}{398--417}.
\newblock


\bibitem[\protect\citeauthoryear{Shi and Eberhart}{Shi and Eberhart}{1998}]%
        {shi1998pso}
\bibfield{author}{\bibinfo{person}{Yuhui Shi} {and} \bibinfo{person}{Russell~C
  Eberhart}.} \bibinfo{year}{1998}\natexlab{}.
\newblock \showarticletitle{Parameter selection in particle swarm
  optimization}. In \bibinfo{booktitle}{\emph{International conference on
  evolutionary programming}}. Springer, \bibinfo{pages}{591--600}.
\newblock


\bibitem[\protect\citeauthoryear{Smiti and Elouedi}{Smiti and Elouedi}{2012}]%
        {smiti2012dbscangm}
\bibfield{author}{\bibinfo{person}{Abir Smiti} {and} \bibinfo{person}{Zied
  Elouedi}.} \bibinfo{year}{2012}\natexlab{}.
\newblock \showarticletitle{Dbscan-gm: An improved clustering method based on
  gaussian means and dbscan techniques}. In \bibinfo{booktitle}{\emph{2012 IEEE
  16th international conference on intelligent engineering systems (INES)}}.
  IEEE, \bibinfo{pages}{573--578}.
\newblock


\bibitem[\protect\citeauthoryear{Vaswani, Shazeer, Parmar, Uszkoreit, Jones,
  Gomez, Kaiser, and Polosukhin}{Vaswani et~al\mbox{.}}{2017}]%
        {vaswani2017attention}
\bibfield{author}{\bibinfo{person}{Ashish Vaswani}, \bibinfo{person}{Noam
  Shazeer}, \bibinfo{person}{Niki Parmar}, \bibinfo{person}{Jakob Uszkoreit},
  \bibinfo{person}{Llion Jones}, \bibinfo{person}{Aidan~N Gomez},
  \bibinfo{person}{{\L}ukasz Kaiser}, {and} \bibinfo{person}{Illia
  Polosukhin}.} \bibinfo{year}{2017}\natexlab{}.
\newblock \showarticletitle{Attention is all you need}.
\newblock \bibinfo{journal}{\emph{Advances in neural information processing
  systems}}  \bibinfo{volume}{30} (\bibinfo{year}{2017}).
\newblock


\bibitem[\protect\citeauthoryear{Veenman, Reinders, and Backer}{Veenman
  et~al\mbox{.}}{2002}]%
        {veenman2002d31}
\bibfield{author}{\bibinfo{person}{Cor~J. Veenman}, \bibinfo{person}{Marcel
  J.~T. Reinders}, {and} \bibinfo{person}{Eric Backer}.}
  \bibinfo{year}{2002}\natexlab{}.
\newblock \showarticletitle{A maximum variance cluster algorithm}.
\newblock \bibinfo{journal}{\emph{IEEE Transactions on pattern analysis and
  machine intelligence}} \bibinfo{volume}{24}, \bibinfo{number}{9}
  (\bibinfo{year}{2002}), \bibinfo{pages}{1273--1280}.
\newblock


\bibitem[\protect\citeauthoryear{Vijay and Nanda}{Vijay and Nanda}{2019}]%
        {vijay2019catalogs}
\bibfield{author}{\bibinfo{person}{Rahul~Kumar Vijay} {and}
  \bibinfo{person}{Satyasai~Jagannath Nanda}.} \bibinfo{year}{2019}\natexlab{}.
\newblock \showarticletitle{A Variable $\epsilon$-DBSCAN Algorithm for
  Declustering Earthquake Catalogs}.
\newblock In \bibinfo{booktitle}{\emph{Soft Computing for Problem Solving}}.
  \bibinfo{publisher}{Springer}, \bibinfo{pages}{639--651}.
\newblock


\bibitem[\protect\citeauthoryear{Vinh, Epps, and Bailey}{Vinh
  et~al\mbox{.}}{2010}]%
        {vinh2010ari}
\bibfield{author}{\bibinfo{person}{Nguyen~Xuan Vinh}, \bibinfo{person}{Julien
  Epps}, {and} \bibinfo{person}{James Bailey}.}
  \bibinfo{year}{2010}\natexlab{}.
\newblock \showarticletitle{Information theoretic measures for clusterings
  comparison: Variants, properties, normalization and correction for chance}.
\newblock \bibinfo{journal}{\emph{JMLR}}  \bibinfo{volume}{11}
  (\bibinfo{year}{2010}), \bibinfo{pages}{2837--2854}.
\newblock


\bibitem[\protect\citeauthoryear{Wei et~al\mbox{.}}{Wei et~al\mbox{.}}{2019}]%
        {wei2019milan}
\bibfield{author}{\bibinfo{person}{Jiabin Wei} {et~al\mbox{.}}}
  \bibinfo{year}{2019}\natexlab{}.
\newblock \showarticletitle{Commercial Activity Cluster Recognition with
  Modified DBSCAN Algorithm: A Case Study of Milan}. In
  \bibinfo{booktitle}{\emph{2019 IEEE International Smart Cities Conference
  (ISC2)}}. IEEE, \bibinfo{pages}{228--234}.
\newblock


\bibitem[\protect\citeauthoryear{Yang and Shami}{Yang and Shami}{2020}]%
        {yang2020hyperparameter}
\bibfield{author}{\bibinfo{person}{Li Yang} {and} \bibinfo{person}{Abdallah
  Shami}.} \bibinfo{year}{2020}\natexlab{}.
\newblock \showarticletitle{On hyperparameter optimization of machine learning
  algorithms: Theory and practice}.
\newblock \bibinfo{journal}{\emph{Neurocomputing}}  \bibinfo{volume}{415}
  (\bibinfo{year}{2020}), \bibinfo{pages}{295--316}.
\newblock


\bibitem[\protect\citeauthoryear{Yang, Lian, Li, Chen, and Li}{Yang
  et~al\mbox{.}}{2014}]%
        {yang2014suspicious_transactions}
\bibfield{author}{\bibinfo{person}{Yan Yang}, \bibinfo{person}{Bin Lian},
  \bibinfo{person}{Lian Li}, \bibinfo{person}{Chen Chen}, {and}
  \bibinfo{person}{Pu Li}.} \bibinfo{year}{2014}\natexlab{}.
\newblock \showarticletitle{DBSCAN clustering algorithm applied to identify
  suspicious financial transactions}. In \bibinfo{booktitle}{\emph{2014
  International Conference on Cyber-Enabled Distributed Computing and Knowledge
  Discovery}}. IEEE, \bibinfo{pages}{60--65}.
\newblock


\bibitem[\protect\citeauthoryear{Zahn}{Zahn}{1971}]%
        {zahn1971compound}
\bibfield{author}{\bibinfo{person}{Charles~T Zahn}.}
  \bibinfo{year}{1971}\natexlab{}.
\newblock \showarticletitle{Graph-theoretical methods for detecting and
  describing gestalt clusters}.
\newblock \bibinfo{journal}{\emph{IEEE Transactions on computers}}
  \bibinfo{volume}{100}, \bibinfo{number}{1} (\bibinfo{year}{1971}),
  \bibinfo{pages}{68--86}.
\newblock


\bibitem[\protect\citeauthoryear{Zheng, Li, Qiu, and Gong}{Zheng
  et~al\mbox{.}}{2012}]%
        {zheng2012maze}
\bibfield{author}{\bibinfo{person}{Kun Zheng}, \bibinfo{person}{Husheng Li},
  \bibinfo{person}{Robert~C Qiu}, {and} \bibinfo{person}{Shuping Gong}.}
  \bibinfo{year}{2012}\natexlab{}.
\newblock \showarticletitle{Multi-objective reinforcement learning based
  routing in cognitive radio networks: Walking in a random maze}. In
  \bibinfo{booktitle}{\emph{2012 international conference on computing,
  networking and communications (ICNC)}}. IEEE, \bibinfo{pages}{359--363}.
\newblock


\bibitem[\protect\citeauthoryear{Zhou and Gao}{Zhou and Gao}{2014}]%
        {zhou2014silhouette_coefficient}
\bibfield{author}{\bibinfo{person}{Hong~Bo Zhou} {and} \bibinfo{person}{Jun~Tao
  Gao}.} \bibinfo{year}{2014}\natexlab{}.
\newblock \showarticletitle{Automatic method for determining cluster number
  based on silhouette coefficient}. In \bibinfo{booktitle}{\emph{Advanced
  Materials Research}}, Vol.~\bibinfo{volume}{951}. Trans Tech Publ,
  \bibinfo{pages}{227--230}.
\newblock


\bibitem[\protect\citeauthoryear{Zhu}{Zhu}{2010}]%
        {zhu2010stream_dataset}
\bibfield{author}{\bibinfo{person}{X. Zhu}.} \bibinfo{year}{2010}\natexlab{}.
\newblock \bibinfo{title}{Stream Data Mining Repository}.
\newblock
\newblock
\urldef\tempurl%
\url{http://www.cse.fau.edu/~xqzhu/stream.html}
\showURL{%
\tempurl}


\end{thebibliography}
